\documentclass[3p, times, 11pt, numafflabel]{elsarticle}
\pdfoutput=1
\usepackage[utf8]{inputenc}
\usepackage[T1]{fontenc}
\usepackage[indent, skip=0.12\baselineskip plus 2pt]{parskip}
\usepackage{setspace, subcaption}
\biboptions{sort}

\journal{Comput. Math. Math. Phys.}

\usepackage{amsmath, babel, booktabs}
\renewcommand{\vec}[1]{\boldsymbol{#1}}
\usepackage[babel=true]{microtype}
\usepackage[group-digits=integer, group-minimum-digits=4]{siunitx}
\usepackage{algorithm, algpseudocode} 
\usepackage{hyperref}
\usepackage[capitalize]{cleveref} 

\newtheorem{theorem}{Theorem}

\newdefinition{definition}{Definition}
\newdefinition{remark}{Definition}
\newproof{proof}{Proof}
\newproof{myproof}{Proof of \cref{thm:1}}

\begin{document}

\begin{frontmatter}

\date{\today}

\title{Global Parameterization-based Texture Space Optimization}

\author[A1,A2]{Wei Chen}

\author[A1,A2]{Yuxue Ren}

\author[A2,A3]{Na Lei\corref{mycorrespondingauthor}}
\cortext[mycorrespondingauthor]{Corresponding author}
\ead{nalei@dlut.edu.cn}

\author[A4]{Zhongxuan Luo}

\author[A5]{Xianfeng Gu}

\address[A1]{Academy for Multidisciplinary Studies, Capital Normal University}
\address[A2]{Beijing Advanced Innovation Center for Imaging Theory and Technology, Capital Normal University}
\address[A3]{DUT-RU ISE, Dalian University of Technology}
\address[A4]{School of Software Technology, Dalian University of Technology}
\address[A5]{State University of New York at Stony Brook, Stony Brook}

\begin{abstract}
Texture mapping is a common technology in the area of computer graphics, it maps the 3D surface space onto the 2D texture space. However, the loose texture space will reduce the efficiency of data storage and GPU memory addressing in the rendering process. Many of the existing methods focus on repacking given textures, but they still suffer from high computational cost and hardly produce a wholly tight texture space.
In this paper, we propose a method to optimize the texture space and produce a new texture mapping which is compact based on global parameterization. The proposed method is computationally robust and efficient. Experiments show the effectiveness of the proposed method and the potency in improving the storage and rendering efficiency.
\end{abstract}


\end{frontmatter}

\section{Introduction}%
\label{sec:introduction}
With rapid development in three-dimensional data acquisition techniques over the last few decades, high resolution model with one or multiple texture images has been produced for a large number of applications. However, many existing model reconstruction methods introduce blank spaces and needless pixels in the attached texture images of the reconstructed models, which make them intractable to perform subsequent storage and computation tasks, such as graphics processing unit (GPU) rendering. 

As one of the geometry processing techniques, the space-oriented texture optimization aims to decrease or remove the blank spaces and the needless pixels of the existing texture images so that an optimized texture space will be generated. Success in space-oriented texture optimization is significantly beneficial to the data storage efficiency and GPU memory addressing in the rendering process. In the research area of optimizing textures of 3D models, we have witnessed a variety of methods over the past decades. Most of these methods optimize the texture space by stretching, shrinking and packing parts of a texture image~\cite{balmelli2002space,martinez2010space,levy2002least,gu2002geometry,sander2003multi}. However, a few limitations exist in these methods, which lie in either incomplete utilization ratio which cannot take full use of texture space, or multiple texture atlases which reduce the efficiency of GPU memory addressing. In some extreme situations, unguaranteed computation of texture mapping due to poor quality of models' triangulation is also a restriction for the application. 

In more recent years, 3D modeling has attracted broad attention from the 3D computer vision community due to the popularity of virtual reality (VR), augmented reality (AR) and metaverse. And photogrammetry is one of the widely used technologies in the field of 3D modeling, it usually produces sophisticated or large scene 3D models with multiple textures. When the data is imported into the VR, AR or metaverse platforms, the real-time performance becomes crucial and important. Therefore, in the area of texture optimization of photogrammetry, generating a tight and global texture space via surface parameterization not only benefits the data storage efficiency on the hard disk and the random access memory (RAM), but also enables the improvement of GPU rendering via speeding up the texture addressing. 

In this work, we propose a novel fast and robust space-oriented texture optimization method. The algorithm is detailed in some steps involving computations of hole filling, topological denoising, cut graph, and parameterization. The main contributions in this work are as follows: \\
(1). Novel framework. Our proposed algorithm is novel and first applied to the space-oriented texture optimization based on global parameterization. It avoids repacking the atlas charts which is considered a NP-hard problem~\cite{book1980michael,milenkovic1999rotational}.\\
(2). Fast and robust computation. Benefit from the fast and robust algorithms of harmonic map and topological denoising, our method is computationally efficient and robust. \\
(3). Efficient result. Compared with existing methods based on texture patches repacking, our algorithm produces one or multiple new textures with no redundant space or pixel.


\section{Related Work}
\label{sec:related_work}
Numerous methods have been proposed for the texture space optimization. Frueh et al.~\cite{frueh2004automated} proposed an approach to packing multiple textures of a model into one single texture atlas in order to optimize the model for rendering purposes, then it simply use a greedy algorithm which copies contiguous texture patches and places them into available space in the atlas.
L{\'e}vy et al.~\cite{levy2002least} designed a surface parameterization method based on least-squares approximation of the Cauchy-Riemann equations. And then they segmented the computed texture atlas into several charts with natural shapes. Thirdly each chart is rescaled to make its area in texture space equals to its area in surface space. At last, before inserting these charts one by one, they are oriented vertically based on the maximum diameter and sorted in decreasing order.
Balmelli et al.~\cite{balmelli2002space} presented a new texture optimization algorithm based on the reduction of the physical space allotted to the texture image, the result image is stretched in high frequency areas and shrunk in low frequency regions, but there still exists redundancy spaces and pixels.
Nöll et al.~\cite{noll2011efficient}  introduced the new search space of modulo valid packings, the key idea was to allow the texture charts to wrap around in the atlas, then based on this search space they proposed an algorithm that can be used to pack texture atlases automatically. 

Recently, Odaker et al.~\cite{odaker2016texture} proposed an approach that aims to remove the redundancies in order to reduce the amount of memory required to store the texture data. In that work, the per-triangle textures are compiled into a new set of texture images of user defined size. The final texture data is packed as tightly as possible.
Limper et al.~\cite{limper2018box} utilized the box cutter technology to refine the texture atlas by eliminating spaces iteratively. This method improves packing efficiency without changing distortion by strategically cutting and repacking the atlas charts. Therefore it preserves the local mapping between the 3D surface and the atlas charts.

All the methods above can be summarized into the problem of how to pack 2D charts efficiently inside the texture rectangle. It is also considered to be a problem of polygon packing which is a combinatorial optimization problem in the field of computational geometry. However, maximizing the packing efficiency for a given set of charts is a NP-hard problem~\cite{book1980michael,milenkovic1999rotational}, thus it can only be deal with empirically.


\section{Theoretic Background}
\label{sec:theory}
This section briefly explains the basic concepts and theorems referred in this paper. 

\begin{figure}[h]
  \centering{}
  \subcaptionbox{\label{fig:surface}}{
    \includegraphics[width=0.3\linewidth]{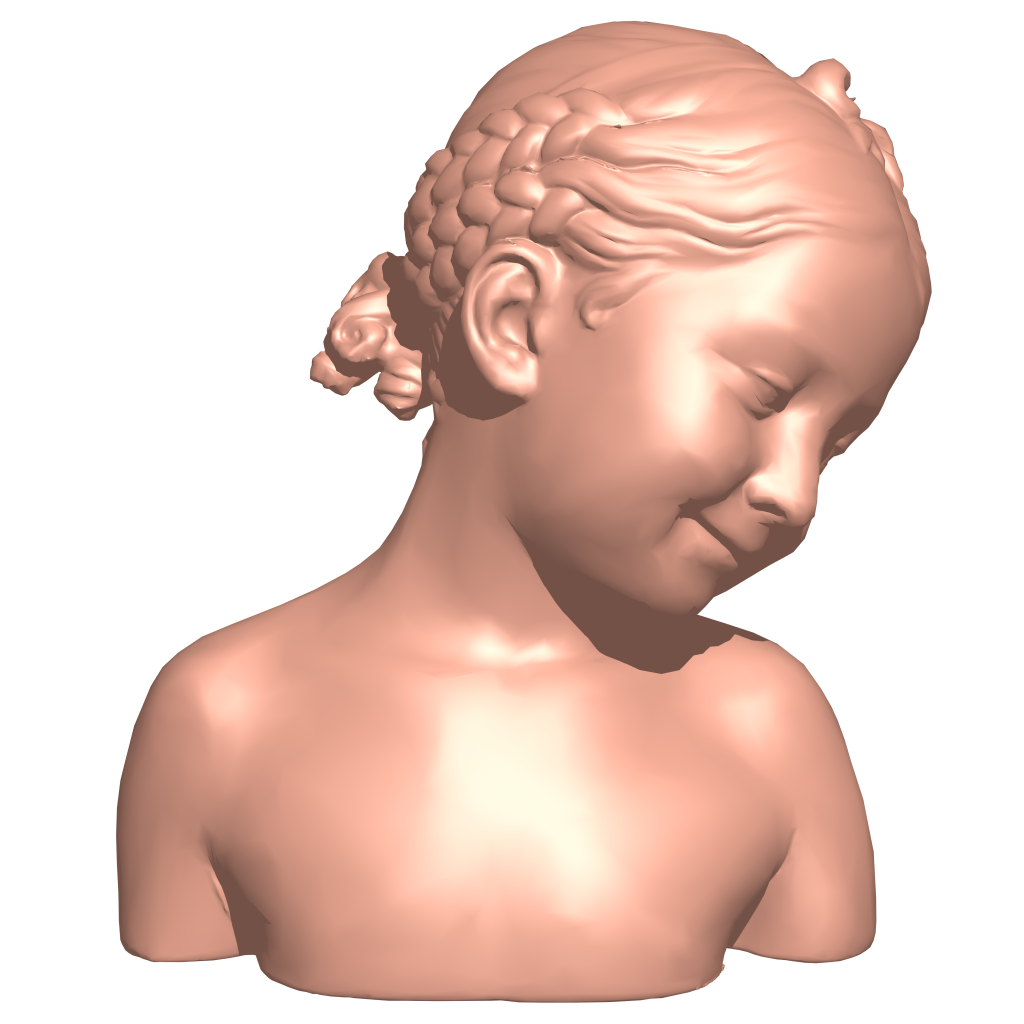}
  }
  \qquad{}
  \subcaptionbox{\label{fig:mesh}}{
    \includegraphics[width=0.3\linewidth]{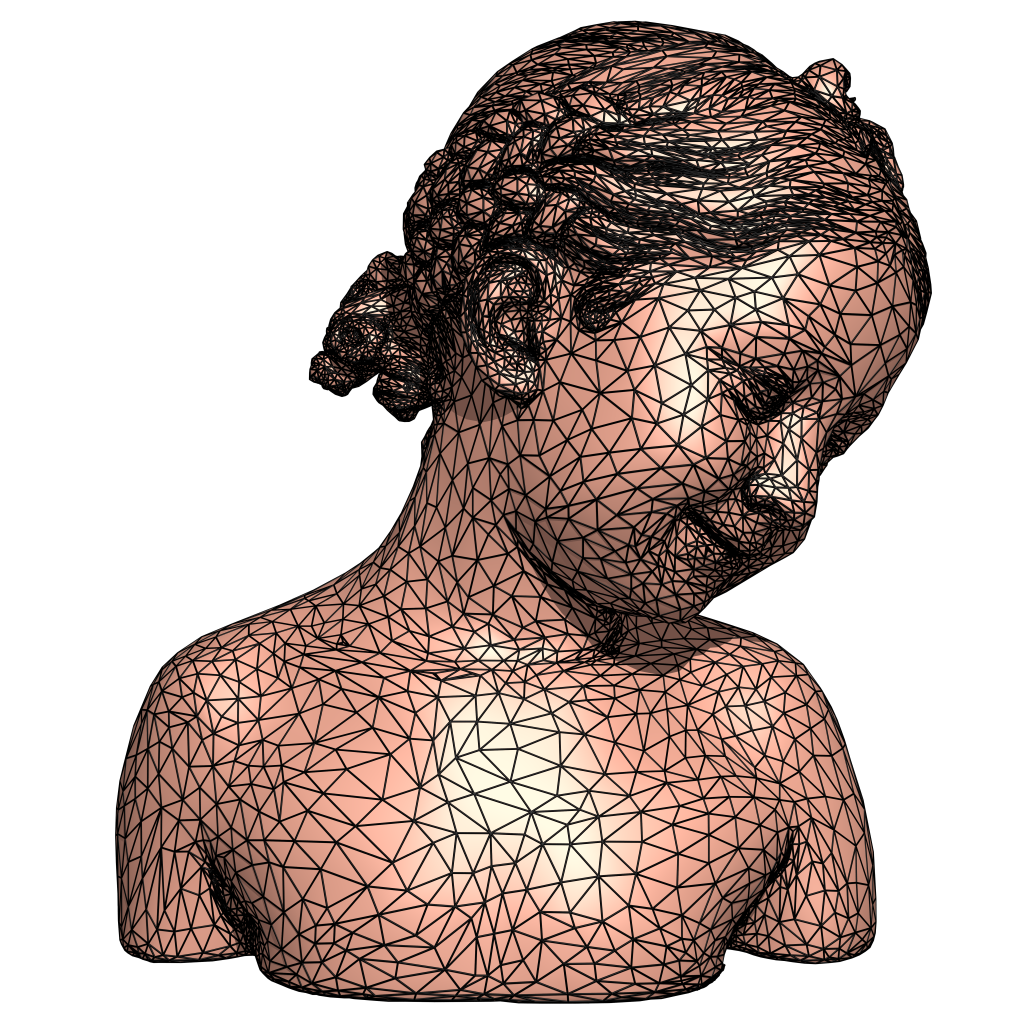}
  }
  \caption{
    Smooth surface and its discretization:
    \subref{fig:surface} a smooth surface of a girl, \subref{fig:mesh} a discretization version of the surface which is a polygon mesh here.}
  \label{fig:surface_mesh}
\end{figure}

Suppose $S$ is a smooth surface, and polygonal mesh $M$ is a discretization of the surface $S$, \cref{fig:surface_mesh} gives an example of a smooth surface and a polygonal mesh.

\begin{definition}[Genus] Given a surface $S$, the number of its handles is called the \emph{genus} of the surface.
\end{definition}

\begin{figure}[h]
  \centering{}
  \subcaptionbox{\label{fig:genus_zero}}{
    \includegraphics[height=0.18\textheight]{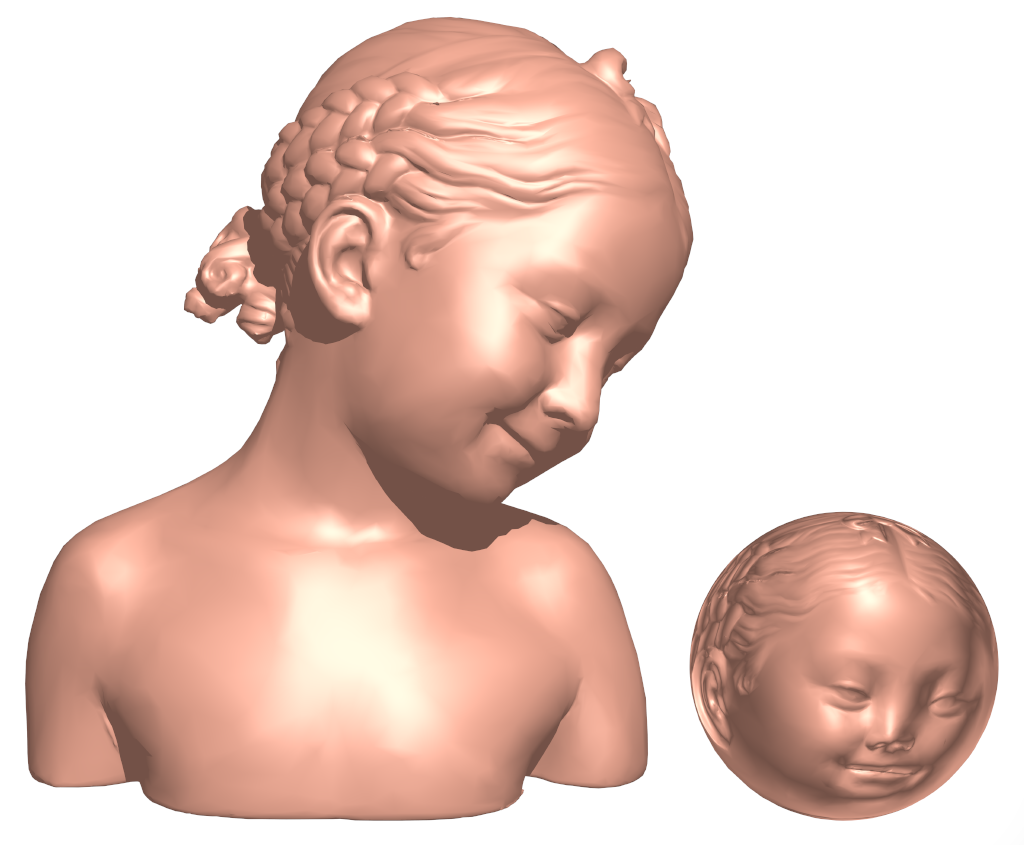}
  }
  \quad{}
  \subcaptionbox{\label{fig:genus_one}}{
    \includegraphics[height=0.18\textheight]{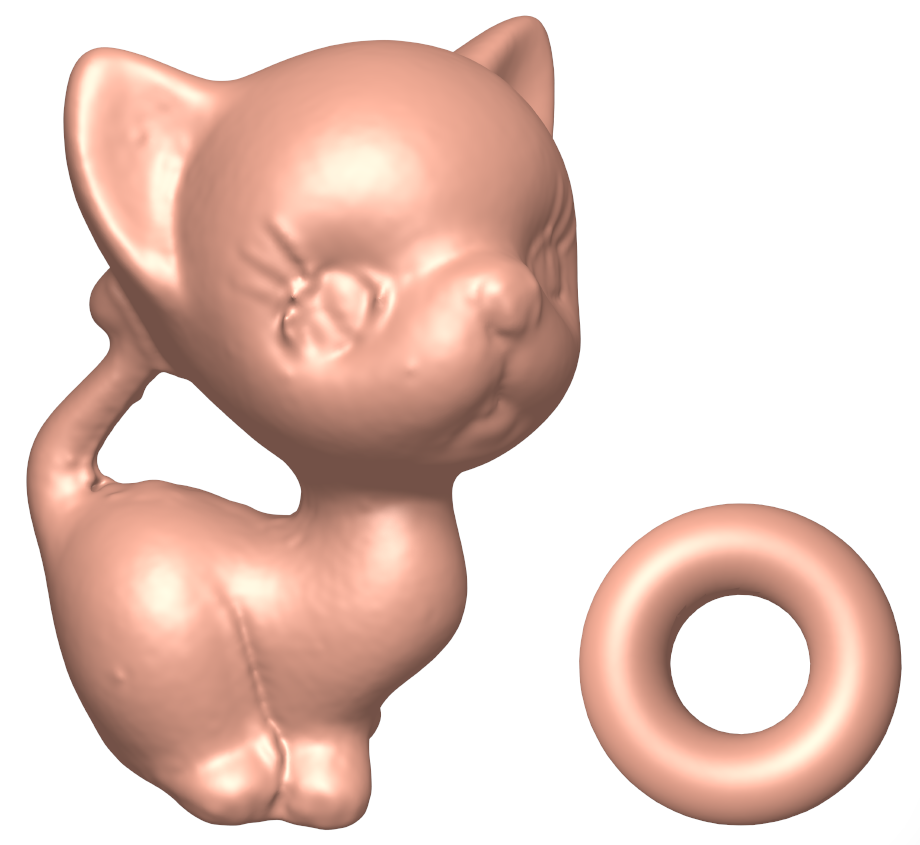}
  }
  \quad{}
  \subcaptionbox{\label{fig:genus_two}}{
    \includegraphics[height=0.18\textheight]{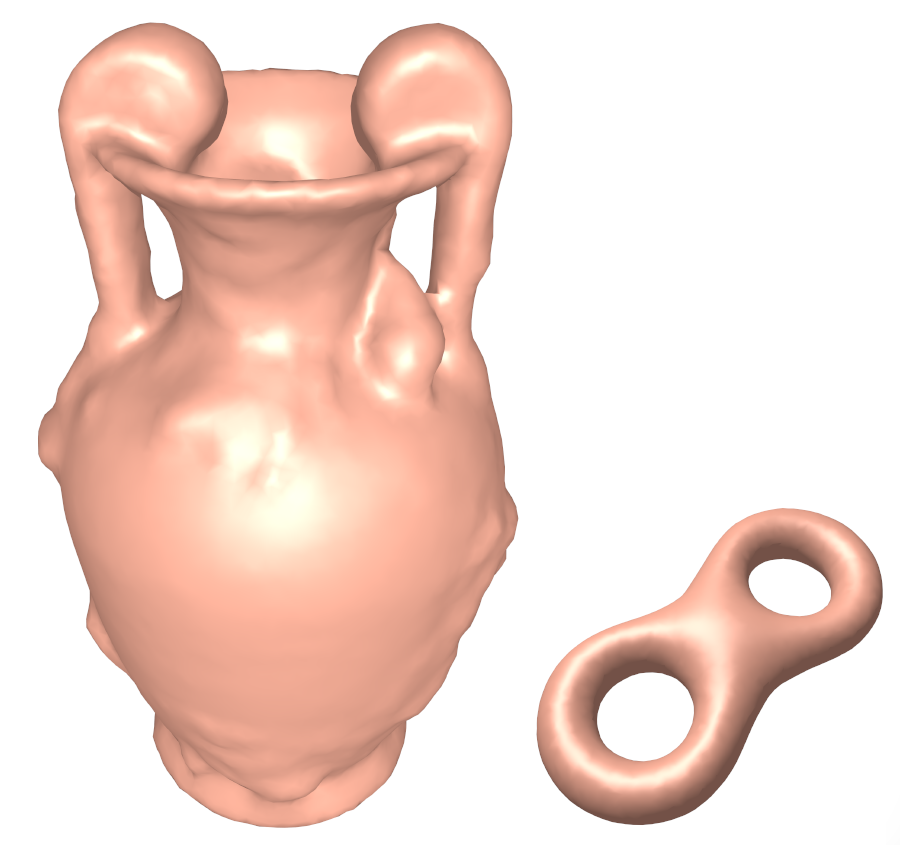}
  }
  \caption{
    Surfaces with different genus:
    \subref{fig:genus_zero} genus zero surfaces, \subref{fig:genus_one} genus one surfaces, and \subref{fig:genus_two} genus two surfaces. The two surfaces in each picture have the same genus.}
  \label{fig:genus}
\end{figure}

The surface's genus is the major topological invariant. We can imagine that the surface is made of elastic rubber, it can be stretched and compressed but not be tore up. Topology only consider the global properties of a space. For example, there is a model in the left picture in \cref{fig:genus}, it can be deformed elastically to a unit sphere surface without tearing the surface.

Given any point $p$ on surface $S$, if the local neighbourhood of $p$ is always homeomorphous to a unit disk, then $S$ is a \emph{closed} surface. Or simply speaking, a surface is closed if there is no boundary on it. However, not all of the surface are closed.

\begin{definition}[Boundary] Suppose surface $S$ is a subset of a topological space $X$, the \emph{boundary} of the surface $\partial S$ is the closure of $S$ minus the interior of $S$ in $X$.
\end{definition}

\begin{figure}[h]
  \centering{}
  \subcaptionbox{\label{fig:cg_eight}}{
    \includegraphics[width=0.3\linewidth]{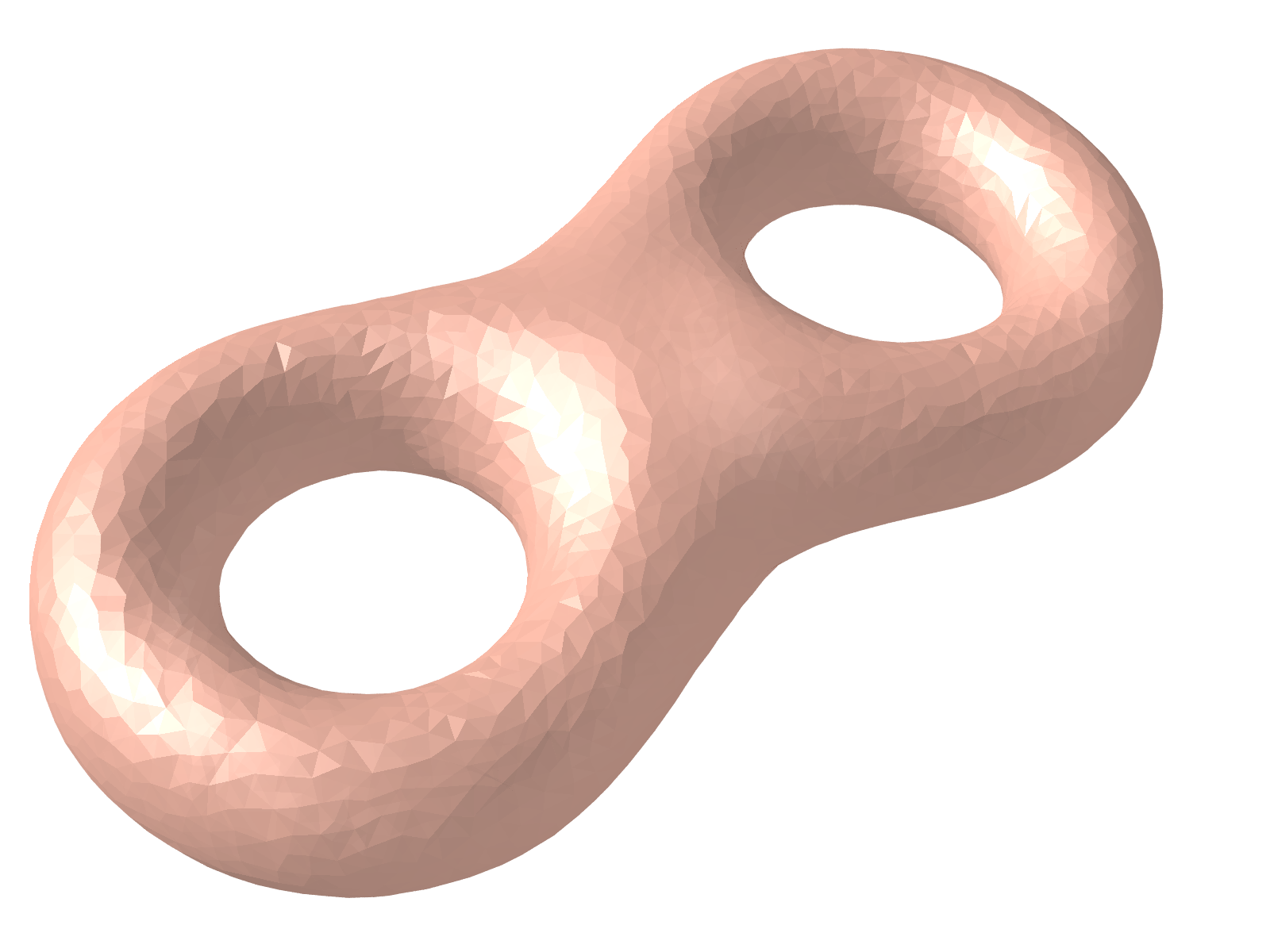}
  }
  \qquad{}
  \subcaptionbox{\label{fig:cg_slice}}{
    \includegraphics[width=0.3\linewidth]{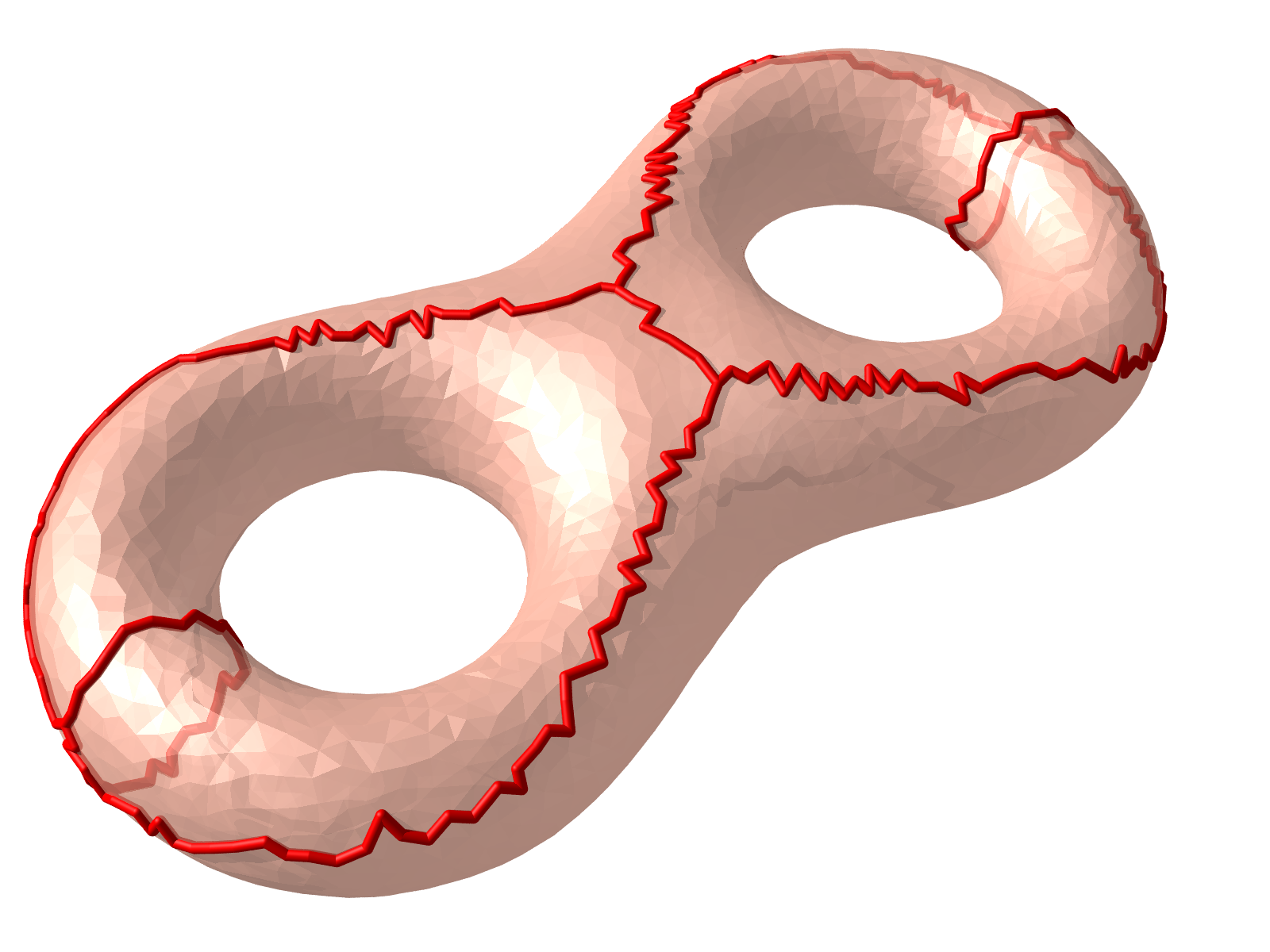}
  }
  \caption{
    Cut graph of a mesh:
    \subref{fig:cg_eight} an input mesh, \subref{fig:cg_slice} the fundamental domain produced by slicing along the cut graph in red.}
  \label{fig:cut_graph}
\end{figure}

A surface with boundaries is called an \emph{open} surface. The number of boundaries is also a topological invariant. In \cref{fig:cg_slice}, the surface is sliced by its cut graph, then it obviously has one boundary, and is topologically equivalent to a unit disk.

\begin{definition}[Cut Graph] Given a closed mesh $M$, a \emph{cut graph} is a set of edges ${e_i}$, which could cut the mesh into a topological disk.
\end{definition}

If a closed mesh is sliced by its cut graph, the result mesh is called its \emph{fundamental domain}. For example, in \cref{fig:cut_graph} the right image shows the fundamental domain of the input mesh.

Let $S$ be a simply connected, closed and orientable surface, embedded in the Euclidean space $\mathbb{R}^3$. We compactify $\mathbb{R}^3$ to a three-dimensional sphere $\mathbb{S}^3$ by adding one infinity point $\infty$, then the open set of the infinity
\begin{equation*}
U_r=\{(x,y,z)|x^2+y^2+z^2>r^2\},
\end{equation*}
where $r$ is the radius of sphere $\mathbb{S}^3$.

Surface $S$ separate $\mathbb{S}^3$ into two parts, the inside part $I$ and the outside part $O$. We define two sets 
\begin{equation*}
I_S = I \cup S, O_S = O \cup S.
\end{equation*}

\begin{definition}[Handle Loop] A loop on $S$ is a \emph{handle loop} if the homology class carried by it is trivial in $I_S$'s homology group $H_1(I_S)$ and non trivial in $O_S$'s homology group $H_1(O_S)$.
\end{definition}

\begin{definition}[Tunnel Loop] A loop on $S$ is a \emph{tunnel loop} if the homology class carried by it is trivial in $H_1(O_S)$ and non trivial in $H_1(I_S)$.
\end{definition}

The following theorem shows that the tunnel and handle loops are unique and form a homology basis~\cite{dey2007computing}.

\begin{theorem}[Handle and Tunel Loops]
\label{thm:handle_tunnel_loops}
For any connected closed surface $S \hookrightarrow \mathbb{S}^3$ embedded in the three-dimensional sphere $\mathbb{S}^3$ of genus $g$, there exists $g$ handle loops $\{h_1, h_2, ..., h_g\}$ forming a basis for $H_1(O_S)$ and $g$ tunnel loops $\{t_1, t_2, ..., t_g\}$ forming a basis for $H_1(I_S)$. Furthermore, $\{h_1, h_2, ..., h_g\}$ and $\{t_1, t_2, ..., t_g\}$ form a basis for $H_1(S)$.
\end{theorem}

The first proof is given in the work~\cite{dey2007computing} by Dey et al. 

\begin{figure}[h]
  \centering{}
  \subcaptionbox{\label{fig:genus3}}{
    \includegraphics[width=0.3\linewidth]{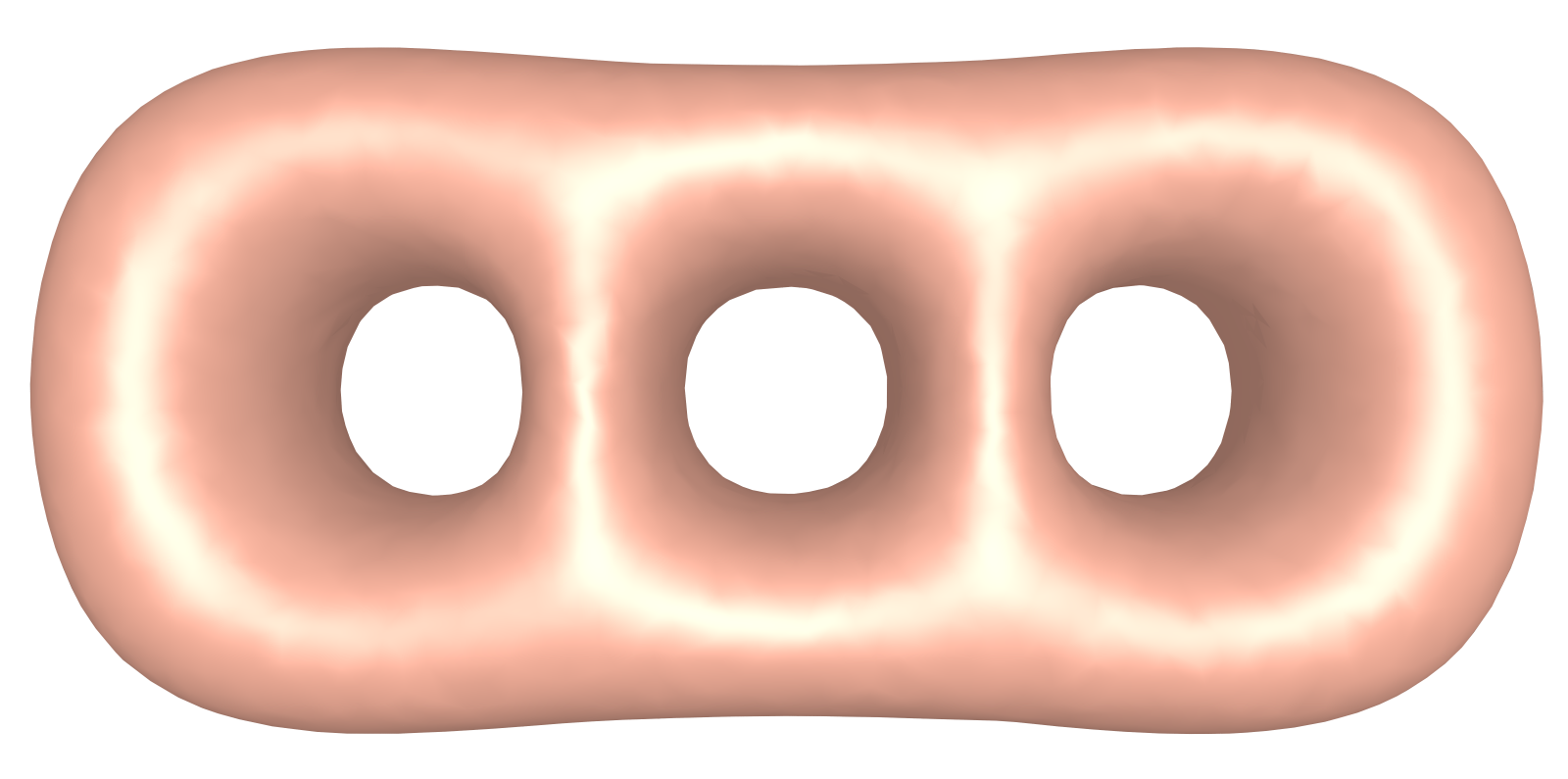}
  }
  \qquad{}
  \subcaptionbox{\label{fig:genus3_loops}}{
    \includegraphics[width=0.3\linewidth]{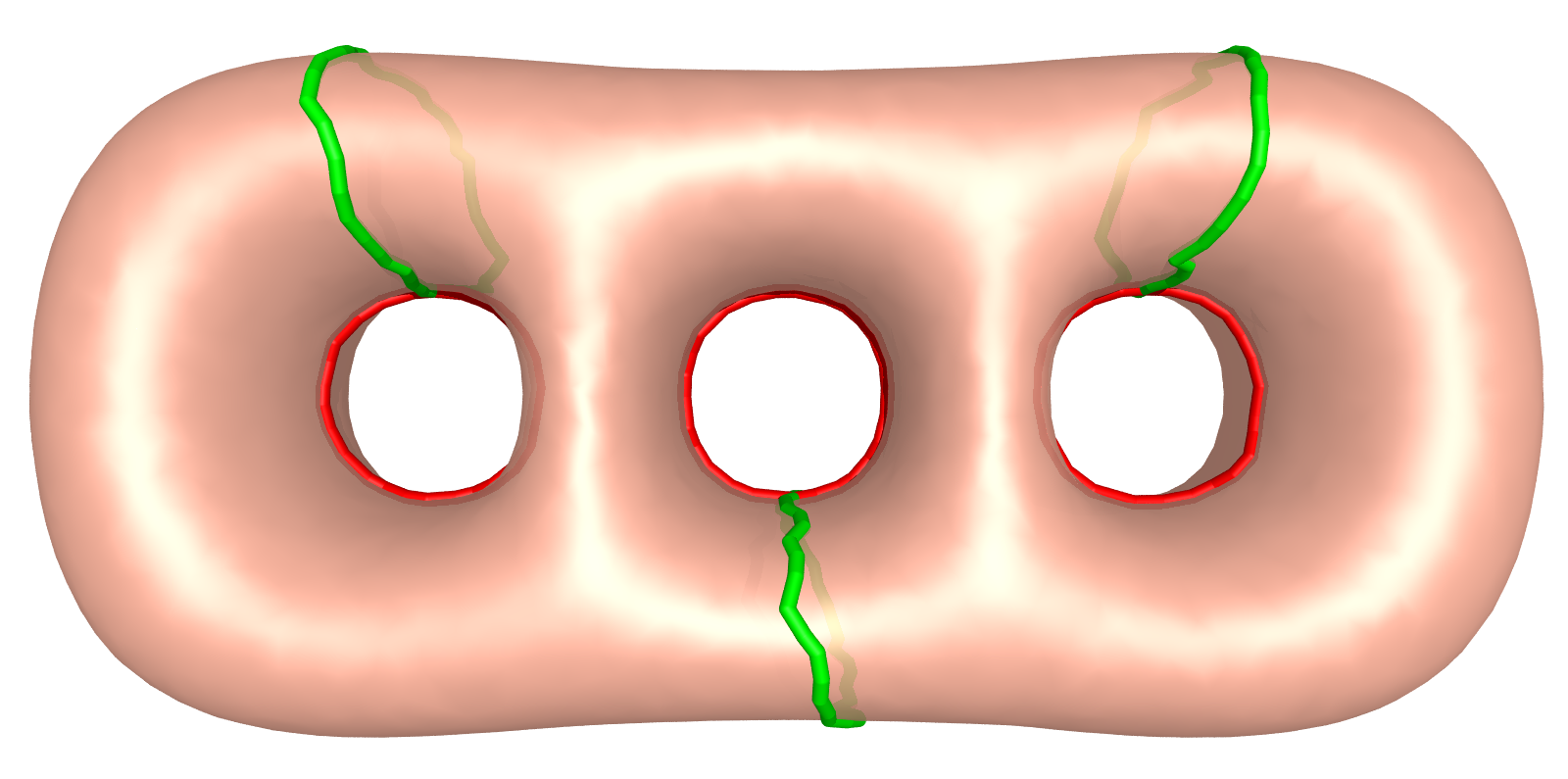}
  }
  \caption{
    Handle and tunnel loops:
    \subref{fig:genus3} a closed surface of genus three, \subref{fig:genus3_loops} the green lines show the handle loops, and the red lines show the tunnel loops.}
  \label{fig:homology_basis}
\end{figure}

For instance, in~\cref{fig:homology_basis}, the left figure is a genus three surface, it has three handle loops and three tunnel loops which are colored in green and red respectively. All of the six loops form a basis of its homology group.

In differential geometry, a parametric curve is defined as a function from $\mathbb{R}$ to $\mathbb{R}^2$ (plane curve) or $\mathbb{R}$ to $\mathbb{R}^3$ (space curve). This definition can be extended to surface situation.

\begin{definition}[Parameterized Surface] A \emph{parameterized surface} $S: U \subset \mathbb{R}^2 \to \mathbb{R}^3$ is a differentiable map from an open set $U \subset \mathbb{R}^2$ into $\mathbb{R}^3$.
\end{definition}

In this paper, we use harmonic map, which will be described in \cref{subsec:harmonic_map}, to construct a desired parameterized surface.

\section{Proposed Method and Algorithm}
\label{sec:algorithm}
In a general way, the texture space of a 3D model produced by the technology of photogrammetry is divided into several parts. Also, there are a mess of blank spaces between these parted domains.

In this section, we will give an overview framework for optimizing the texture space first, and then the details of each step will be discussed separately.

Since the pipeline of 3D modeling in photogrammetry is mainly divided into two steps: image registration and texture selection. This method hardly avoids the fragmentation of texture space. We propose a novel method to compute a new texture mapping and corresponding texture image, it will fully eliminate the blank space in the texture space and unnecessary pixels. 

\subsection{Data Preparation}
\label{subsec:data_preparation}

In the step of data preparation, a model of topological disk is needed for computing a global parameterization. Therefore there are three sub-steps for data preparation. The first one is to fix the non-manifold elements. The second one is hole filling. The last one, which is also the most difficult part of the three, is topological denoising. 

There maybe exists non-manifold elements in the input 3D model, but this can be worked out by simply slicing the non-manifold vertices and edges. If the model is sliced into two or more parts, then we process them separately.

Sometimes a few holes, most of them are small, also exist on the models, they can be filled directly. Suppose the hole or boundary is denoted as $B=\{e_0, e_1, ..., e_{n-1}\}$, it consists of $n$ edges, each edge $e_i=(v_i, v_{i+1}), i=0,1,...n-1, v_0=v_n$, so there are also $n$ vertices on the boundary. We create a new vertex $v_{new}$ which its coordinate equals to $\frac{1}{n} \sum_{i=0}^{n-1} coord(v_i)$, then $n$ new faces $f_i=(v_{new}, v_i, v_{i+1})$ are inserted into the model to close one hole.

\begin{figure}[h]
  \centering{}
  \subcaptionbox{\label{fig:eight}}{
    \includegraphics[width=0.3\linewidth]{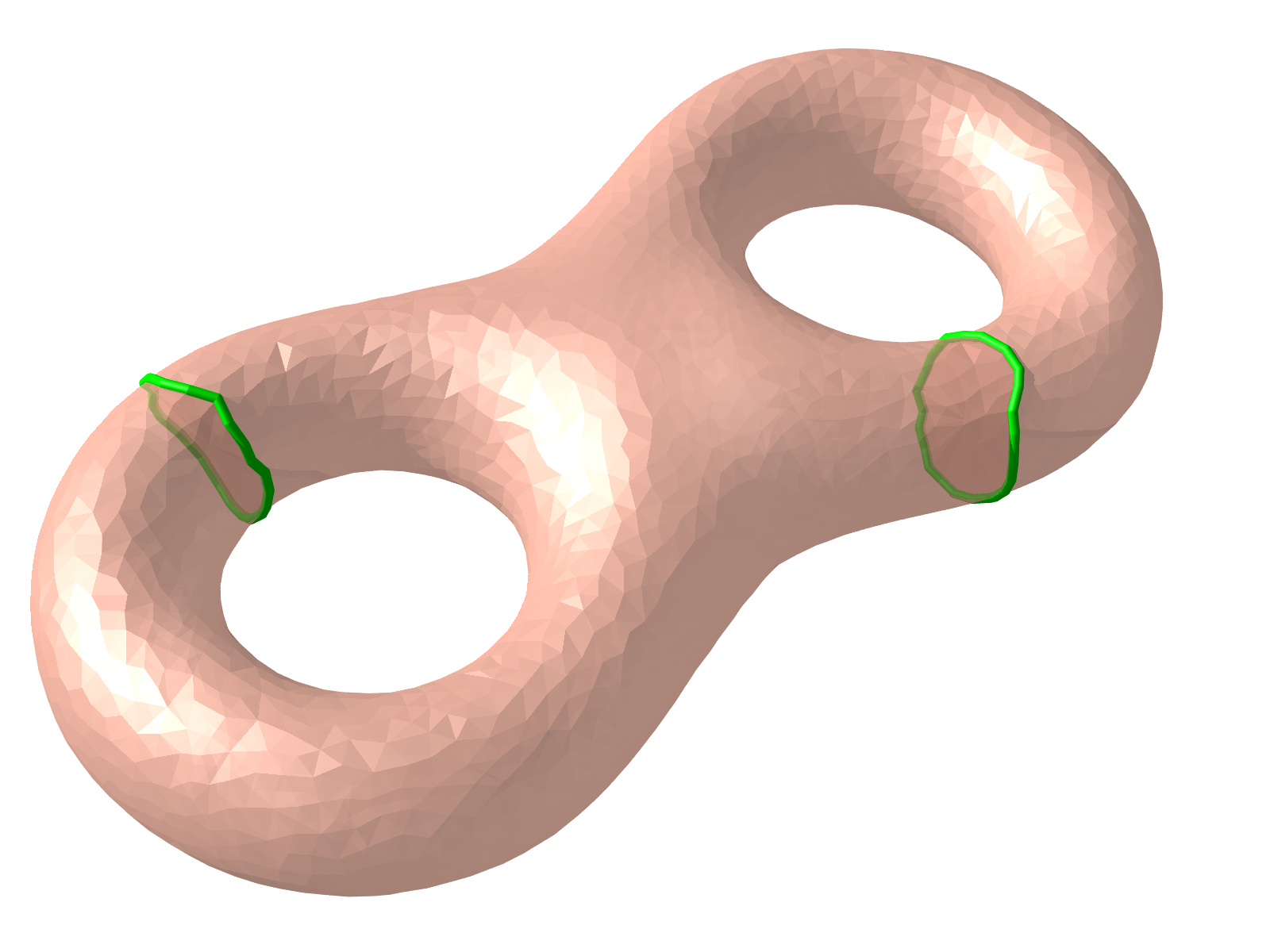}
  }
  \qquad{}
  \subcaptionbox{\label{fig:clifford}}{
    \includegraphics[width=0.3\linewidth]{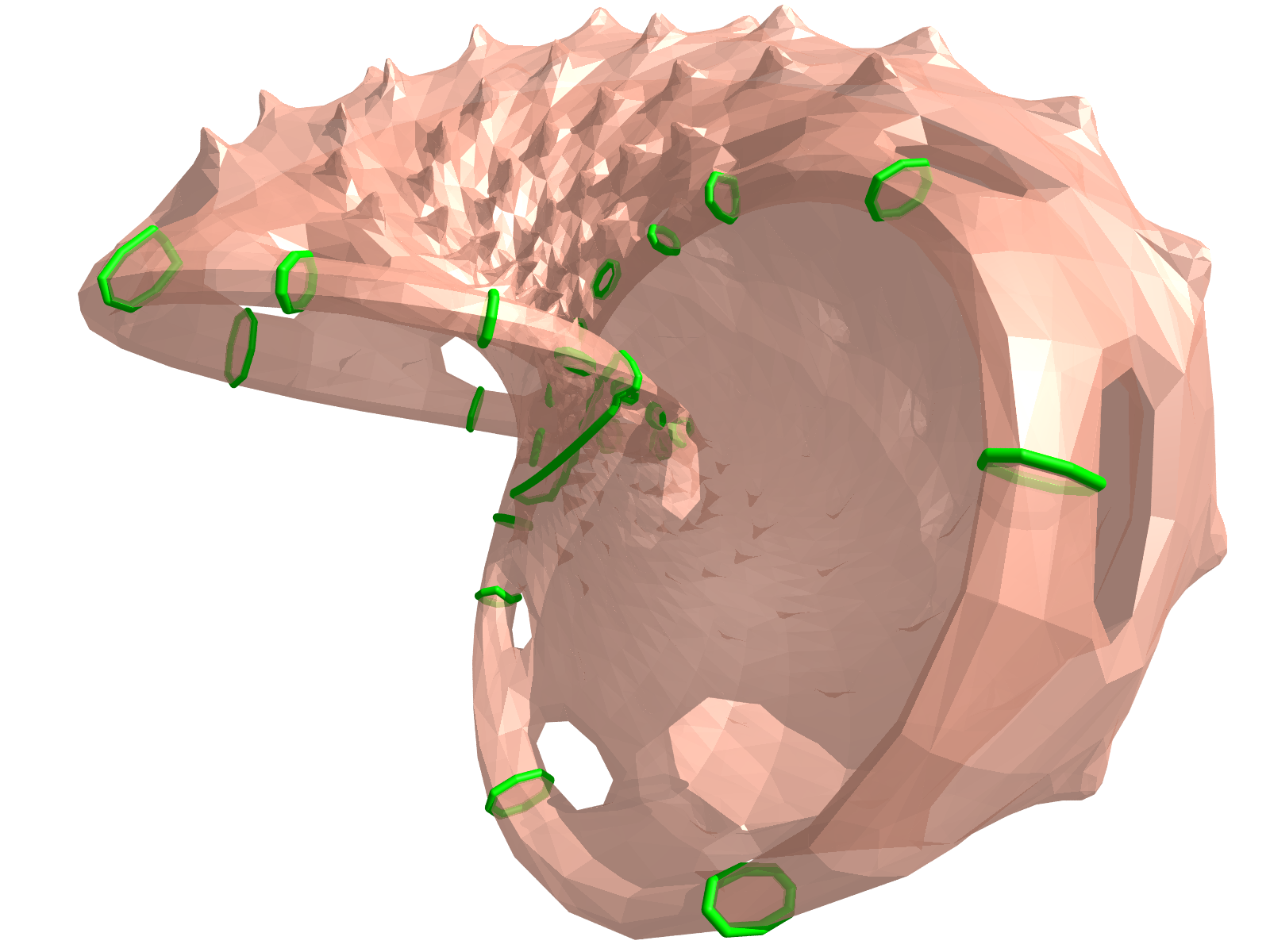}
  }
  \caption{
    Topological denoising based on homology detection, the green loops are the handle or tunnel loops, which their number is equal to the genus of the surface.
    \subref{fig:eight} a simple surface with genus $2$, \subref{fig:clifford} a complicated surface with genus $25$.}
  \label{fig:topological_noise}
\end{figure}

After eliminating the non-manifold elements and filling the small holes, the input mesh becomes a closed mesh with genus $g$. Usually, these minor handles are topological noises, there are several methods to detect them, some of the methods are based on homotopy and homology~\cite{erickson2005greedy,chen2008quantifying,dey2008computing}, and some are based on graph structures such as Reeb graph, core graph, medical axis or curve skeletons~\cite{dey2013efficient,dey2007computing,dey2009computing,erickson2012combinatorial,zhou2007topology}. In our work, we use the homology based method to remove the topological noises since its robustness and efficiency compared with other methods. \cref{fig:topological_noise} shows two examples for detecting the handles.

\subsection{Global Parameterization Based on Harmonic Map}
\label{subsec:harmonic_map}

A harmonic map $H: M \to D$ can be considered as an planar embedding from a space surface $M$ of topological disk to a planar domain $D$, where $D$ is a convex domain, and in our paper, we always consider $D$ is a unit square. A harmonic map is always the extreme point for the harmonic energy function, 
\begin{equation*}
E(H)=\int_M|dH|^2d\mu_M,
\end{equation*}
where the norm of differential $dH$ is given by the metric on $M$ and $D$, and $d\mu_M$ is the measure on $M$~\cite{eells1964harmonic,eck1995multiresolution,schoen1994lectures,o2006elementary}. $H$ can be solved by minimizing $E(H)$. 

In order to solve the harmonic map $H$, $H$ can be represented as a pair of functions $(H_1, H_2)$, where $H_i:M\to \mathbb{R}, i\in\{1,2\}$. In this paper, $(H_1, H_2)$ is the parametric coordinate $(u, v)$ in the unit square $D$. Therefore, the harmonic energy can be expressed as 
\begin{equation*}
E(H)=\int_M(|\nabla H_1|^2 + |\nabla H_2|^2).
\label{eq:harmonic_energy}
\end{equation*}
To minimizing the harmonic energy, a harmonic map can be solved by the Euler-Lagrange differential equation of the energy functional, i.e. $\Delta H=0$, where $\Delta$ is the Laplace-Beltrami operator~\cite{eells1964harmonic,eck1995multiresolution,schoen1994lectures,o2006elementary}.

\begin{figure}[h]%
  \centering{}%
  \includegraphics[width=0.26\linewidth]{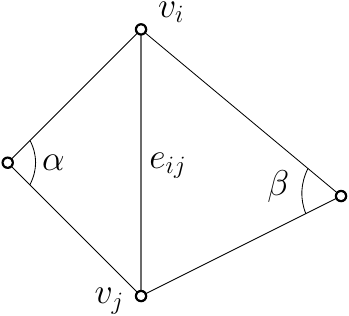}%
  \caption{Illustration the two angles $\alpha$ and $\beta$ against the edge $e_{ij}$.}%
  \label{fig:cot_weight}
\end{figure}

In discrete case, the source domain $M$ is represented as a triangle mesh, the discrete harmonic energy can be approximated~\cite{eck1995multiresolution,zhang1999harmonic,gu2003surface} as
\begin{equation}
E(H)=\sum_{e_{ij}\in M} w_{e_{ij}}|H(v_1)-H(v_0)|^2,
\end{equation}
where $e_{ij}=[v_i, v_j]$ is an edge with two endpoints $v_i$ and $v_j$, and $w_{ij}$, the edge weight defined on the edge $e_{ij}$, is defined as
\begin{equation*}
k_{e_{ij}}=
\left\{\begin{matrix}
 \frac{1}{2}(\cot{\alpha}+\cot{\beta}), & \forall e_{ij} \notin \partial M \\
 \frac{1}{2}\cot{\alpha}, & otherwise \\
\end{matrix}\right.,
\end{equation*}
where $\alpha$ and $\beta$ are two angles against the edge $e_{ij}$, as shown in \cref{fig:cot_weight}.

In order to minimize a discrete harmonic energy, we also need to solve the Euler-Lagrange differential equation for the energy functional, i.e. $\Delta H=0$, in the discrete situation. This will lead to solving a sparse linear least-square system for the harmonic map $H$ of all vertices on $M$. According to Rad$\acute{\textrm{o}}$'s Theorem~\cite{schoen1994lectures}, if an arbitrary convex target domain is adopted to compute the harmonic map, then the resulting map depends on the boundary in a continuous manner. In other words, if the boundary condition 
\begin{equation*}
H|_{\partial M}: \partial M \to \partial D,
\end{equation*}
is given and $\partial D$ is convex, then the solution exists and is unique~\cite{eck1995multiresolution,zhang1999harmonic,gu2003surface}.

The harmonic energy optimization is a linear problem and can be solved directly using Newton's method. \cref{alg:harmonic_map} gives the details for computing a harmonic map from a topological disk to a planar unit square domain.

\begin{algorithm}[h]
  \caption{Boundary Condition for Harmonic Map}
  \label{alg:boundary_condition}
  \hspace*{\algorithmicindent} \textbf{Input:} The boundary $\partial M$ of mesh $M$, and four boundary vertices $\{c_0, c_1, c_2, c_3\}$ which will be mapped to four corners of the unit square. \\
  \hspace*{\algorithmicindent} \textbf{Output:} A map $\vec{b}$ which maps the boundary $\partial M$ to an unit square $\partial D$.
  \begin{algorithmic}[1]
    \Procedure{SquareBoundary}{$\partial M, \{c_0, c_1, c_2, c_3\}$} 
      \State{} Set the coordinates of $c_0, c_1, c_2, c_3$ to $(0,0),(1,0),(1,1),(0,1)$ respectively.  
      \While{ $k \in \{0, 1, 2, 3\} $} 
        \State{} $s \gets len(c_k,c_{k+1})$, where $len(c_k,c_{k+1})$ represents path length from $c_k$ to $c_{i+k}$ on $\partial M$. \Comment{$c_4=c_0$}
        \State{} Collect a set of vertices between $c_i$ and $c_{i+1}$ in $\partial M$, denoted as $vlist$.
        \While{ $v \in vlist$}
          \State{} $t=len(c_k, v_i)/s$, $\vec{b}(v)=(1-t)\vec{b}(c_k)+t\vec{b}(c_{k+1})$.
        \EndWhile
      \EndWhile
      \State{} \textbf{return} $\vec{b}$
    \EndProcedure{}
  \end{algorithmic}%
\end{algorithm}

\begin{algorithm}[h]
  \caption{Harmonic map for a topological disk}
  \label{alg:harmonic_map}
  \hspace*{\algorithmicindent} \textbf{Input:} A genus zero mesh $M$ with one boundary $\partial M$, and four boundary vertices $\{c_0, c_1, c_2, c_3\}$ which will be mapped to four corners of the unit square. \\
  \hspace*{\algorithmicindent} \textbf{Output:} A harmonic map $H$ which maps the mesh $M$ to an unit square $D$.
  \begin{algorithmic}[1]
    \Procedure{HarmonicMap}{$M, \partial D$} 
      \State{} Traverse the boundary of $M$, store the boundary vertices to a cycle list $\partial M=\{v_0, v_1, ..., v_{m-1}\}$ orderly, where $m$ is the number of boundary vertices.
      \State{} Compute a map $\vec{b}$ which maps the boundary $\partial M$ to the unit square $\partial D$ by \cref{alg:boundary_condition}.  
      \State{} Optimize the harmonic energy \cref{eq:harmonic_energy} using Newton's method with fixed boundary condition. This is equivalent to solving the following linear system with the boundary vertices fixed.
      \begin{equation*}
          \left\{\begin{matrix}
          \Delta H(v_i)   & = & 0,      & \forall v_i \notin \partial M, \\
          H|_{\partial M} & = & \vec{b},& otherwise. \\
          \end{matrix}\right.
      \end{equation*}
      \State{} \textbf{return} $H$.
    \EndProcedure{}
  \end{algorithmic}%
\end{algorithm}

The square harmonic map could be used for the purpose of texture mapping directly, it maps the surface $M$ to an unit square domain $D$ \footnote{In practice, one can get a rectangular texture map by simply stretching an unit square texture map. So, in order to simplify, we only consider the unit square as the target domain in this paper.}, and provides the texture coordinates for texture mapping, but we still need to generate a texture image adapting to the texture coordinates.

\subsection{Texture Image Generating}
\label{subsec:texture_generating}

The texture space of original model is loose, and this feature leads to inefficiency of data storage and GPU memory addressing. In \cref{subsec:harmonic_map}, we have computed a new and global parameterization as the texture coordinates, but the original texture images do not adapt the new texture coordinates, so in this subsection, we will generate a texture image for that. 

The rough idea to generate a new texture image is that we filling colors into the triangles on the new texture space, then the final image is the result. This idea can be simplified to the \emph{polygon filling algorithm}. There are many polygon filling algorithms, such as scan line\cite{pineda1988parallel,nisha2017review}, boundary fill~\cite{kumar2020comparison}, edge fill~\cite{dunlavey1983efficient}, flood fill~\cite{nosal2008flood,kumar2020comparison} and so on. Each method has its own merits and demerits, and all of them can finish our task. In this paper, we will use the scan line algorithm to fill colors into all triangles of the model, then a new texture image is generated.

The scan line filling algorithm works by intersecting each scan line with polygon edges and then fills the segment cut by the polygon. The ~\cref{alg:poly_filling} describes the overview pipeline of the scan line filling algorithm. 

\begin{algorithm}[h]
  \caption{Scan Line Filling Algorithm}
  \label{alg:poly_filling}
  \hspace*{\algorithmicindent} \textbf{Input:} A given planar polygon $f$ with vertices $v_0, v_1, ...$, and each vertex $v_i=(\textrm{u}^{new}_i, \textrm{v}^{new}_i, \textrm{u}^{ori}_i, \textrm{v}^{ori}_i)$ owns its new and original texture coordiantes.\\
  \hspace*{\algorithmicindent} \textbf{Output:} A list of points $pts$ which discretize the segments on scan lines cut by $f$.
  \begin{algorithmic}[1]
    \Procedure{ScanLineFilling}{Polygon $f$} 
      \State{} Find out the minimum $y_{min}$ and maximum $y_{max}$ from the polygon $f$. 
      \State{} Initialize a list of scan lines $\{l_0, l_1, ...\}$ from $y_{min}$ to $y_{max}$.
      \While{ $l_i \in \{l_0, l_1, ...\} $} 
        \State{} Compute the intersections $\{p_0, p_1,...,p_{n_1}\}$ between one scan line $l_i$ and each edge $e$ of polygon $f$, and intersections are sorted in the ascending order of $x$ coordinate. Then we get a group of segments $\{[p_0, p_1], [p_1, p_2], ..., [p_{n-2}, p_{n-1}] \}$. 
      \State{} Fill all the segments which are inside polygon $f$, and discretize the segments by interpolation method based on the new texture coordinates. Meanwhile, the original texture coordinates of the discrete segments can be obtained by the same interpolation coefficient. Collect the discrete points to $pts$.
      \EndWhile
      \State{} \textbf{return} $pts$.
    \EndProcedure{}
  \end{algorithmic}%
\end{algorithm}

\begin{figure}[h]%
  \centering{}%
  \includegraphics[width=0.46\linewidth]{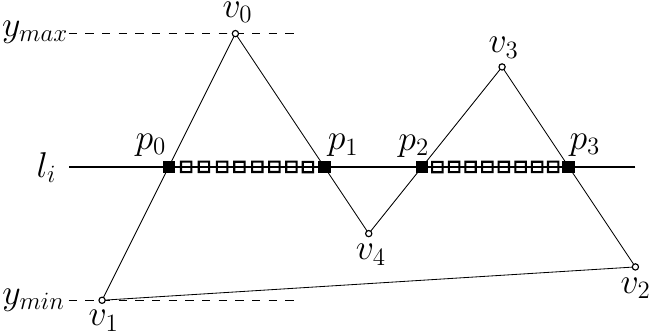}%
  \caption{The scan line $l_i$ intersects polygon $f$ with two segments $[p_0, p_1]$ and $[p_2, p_3]$ which is discretized to several points, here the segment $[p_1, p_2]$ is outside the polygon, so we ignore it.}%
  \label{fig:scan_line}
\end{figure}

A detailed figure is given in ~\cref{fig:scan_line} to explain steps 5 and 6 in ~\cref{alg:poly_filling}. In step 6, the \emph{odd-even rule} could be used to determine whether a segment is inside polygon $f$ or not. In ~\cref{fig:scan_line}, for example, we want to determine whether segment $[p_1, p_2]$ lies inside the polygon or not. Firstly we select any point $p_i$ in $[p_1, p_2]$, then we count the edge crossing along the scan line from point $p_i$ to infinity (either of the two orientations is fine), and the number of intersections is two which is an even number, so segment $[p_1, p_2]$ is outside polygon $f$. If we investigate segments $[p_0, p_1]$ and $[p_2, p_3]$, they both are inside polygon $f$.

\begin{figure}[h]
  \centering{}
  \subcaptionbox{\label{fig:pipeline_1}}{
    \includegraphics[width=0.2\linewidth]{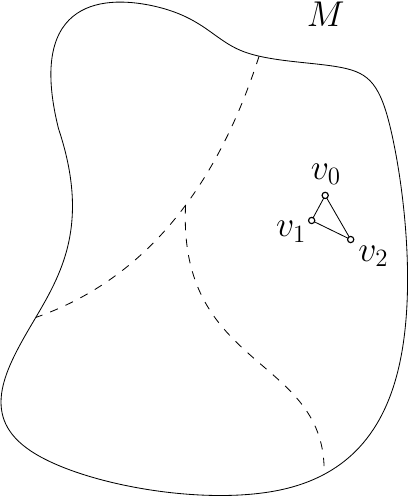}
  }
  \qquad{}
  \subcaptionbox{\label{fig:pipeline_2}}{
    \includegraphics[width=0.25\linewidth]{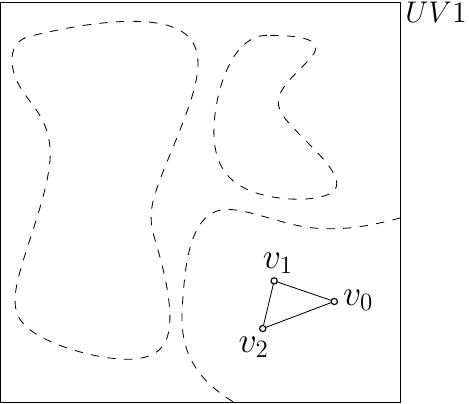}
  }
  \thinspace{}
  \subcaptionbox{\label{fig:pipeline_3}}{
    \includegraphics[width=0.25\linewidth]{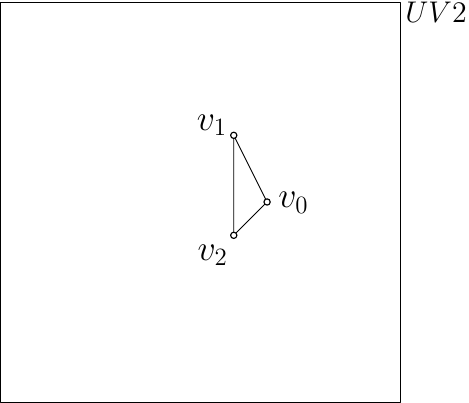}
  }
  \thinspace{}
  \subcaptionbox{\label{fig:pipeline_4}}{
    \includegraphics[width=0.15\linewidth]{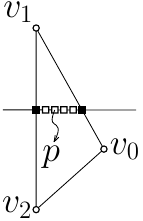}
  }
  \caption{
    Pipeline of generating a new texture image:
    \subref{fig:pipeline_1} a surface $M$ where its texture space is separated into three parts, \subref{fig:pipeline_2} the original fragmented texture space, \subref{fig:pipeline_3} the new texture mapping computed using \cref{alg:harmonic_map} is tight, \subref{fig:pipeline_4} the color of pixel $p$ in $UV2$ domain is achieved via $UV1$ domain, then final texture image is generated.}
  \label{fig:pipeline}
\end{figure}

For each discrete point generated by~\cref{alg:poly_filling}, it has four components $\textrm{u}^{new}, \textrm{v}^{new}, \textrm{u}^{ori}$ and $\textrm{v}^{ori}$, point $p^{new}=(\textrm{u}^{new}, \textrm{v}^{new})$ on new texture space corresponds to point $p^{ori}=(\textrm{u}^{ori}, \textrm{v}^{ori})$ on original texture space. Therefore, the color on $p^{new}$ equals to the color on $p^{ori}$. In this way, a new texture image can be generated efficiently. \cref{fig:pipeline} shows the pipeline of generating a new texture image.

\section{Experiments}
\label{sec:experiments}
All the experiments in this paper are carried out on a Windows laptop with an Intel(R) Core(TM) i7-10710U CPU and a 16GB system memory. The models used in the experiments are created by two types of methods. One is based on structured light scanning system~\cite{zhang2018high}, and the other is the method of multi-view stereo~\cite{moulon2016openmvg,openmvs2020} which uses oblique aerial images as the input. 

\begin{figure}[h]
  \centering{}
  \subcaptionbox{\label{fig:ex1}}{
    \includegraphics[width=0.17\linewidth]{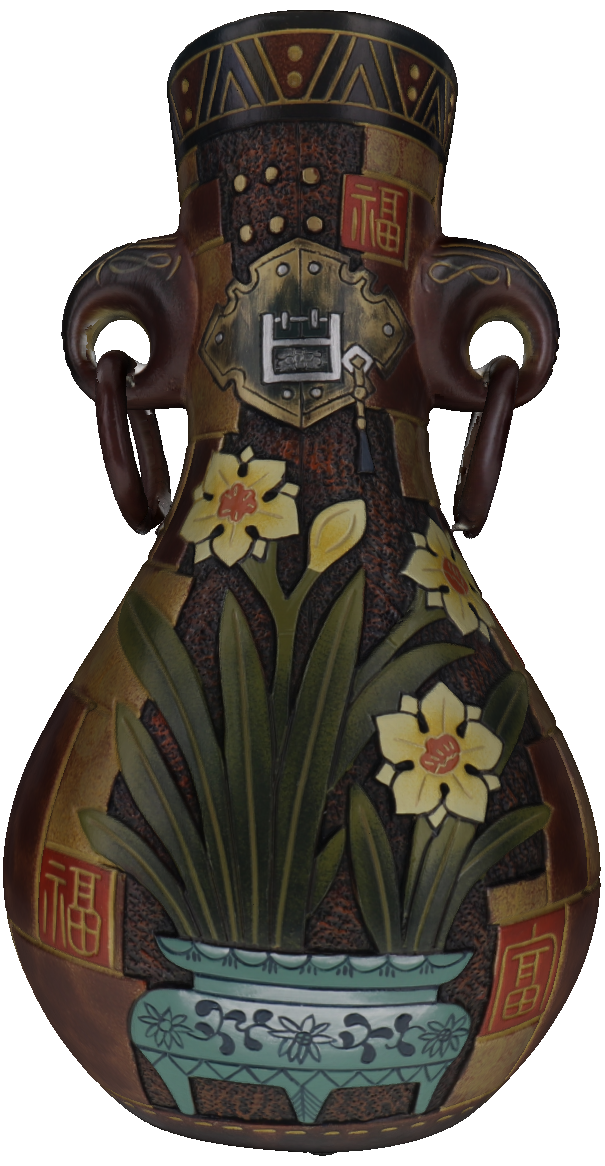}
  }
  \thinspace
  \subcaptionbox{\label{fig:ex2}}{
    \includegraphics[width=0.17\linewidth]{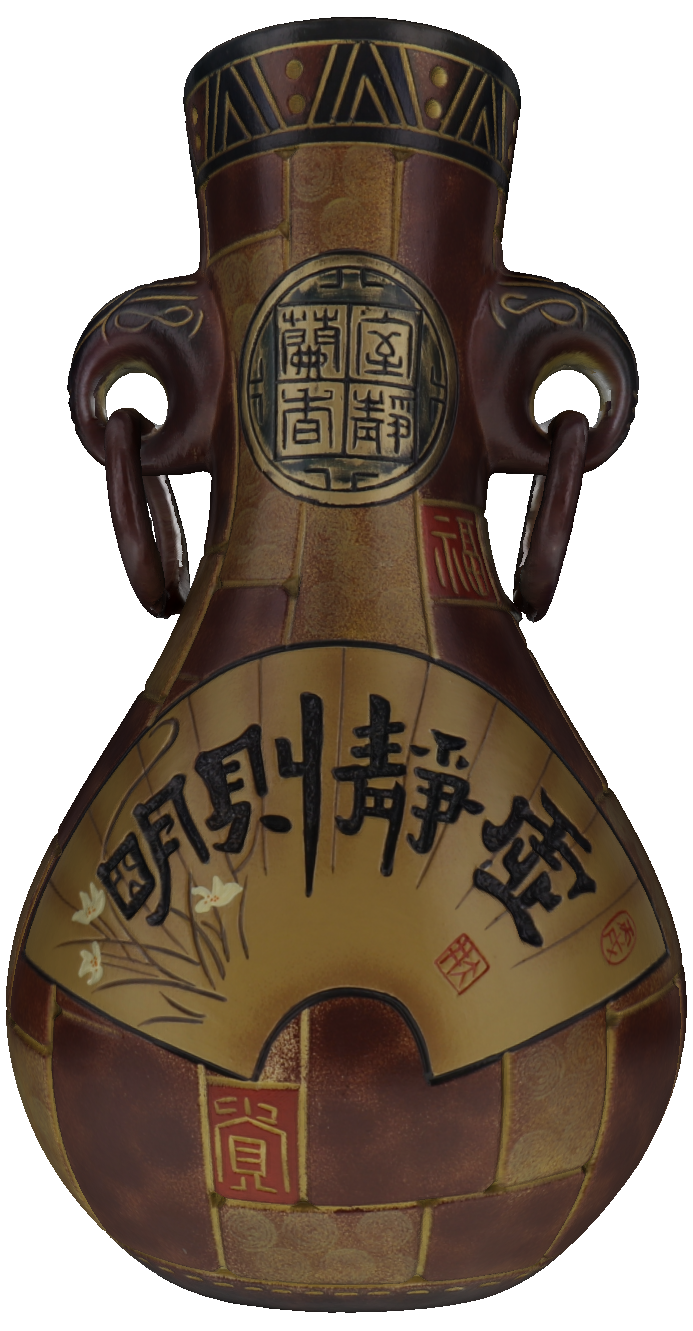}
  }
  \thinspace
  \subcaptionbox{\label{fig:ex3}}{
    \includegraphics[width=0.17\linewidth]{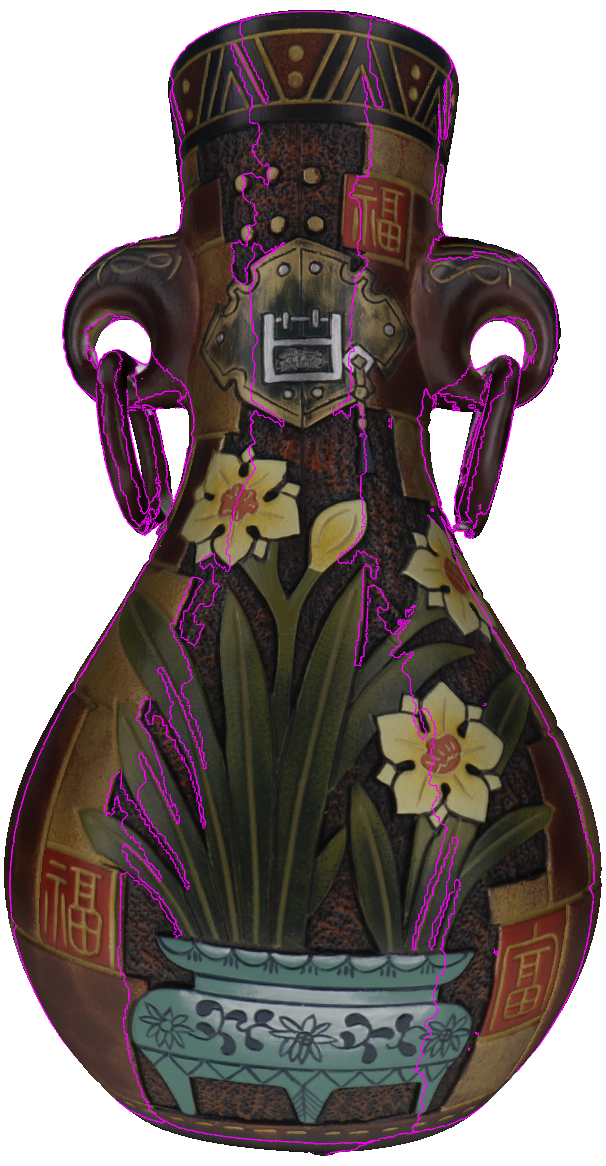}
  }
  \thinspace
  \subcaptionbox{\label{fig:ex4}}{
    \includegraphics[width=0.17\linewidth]{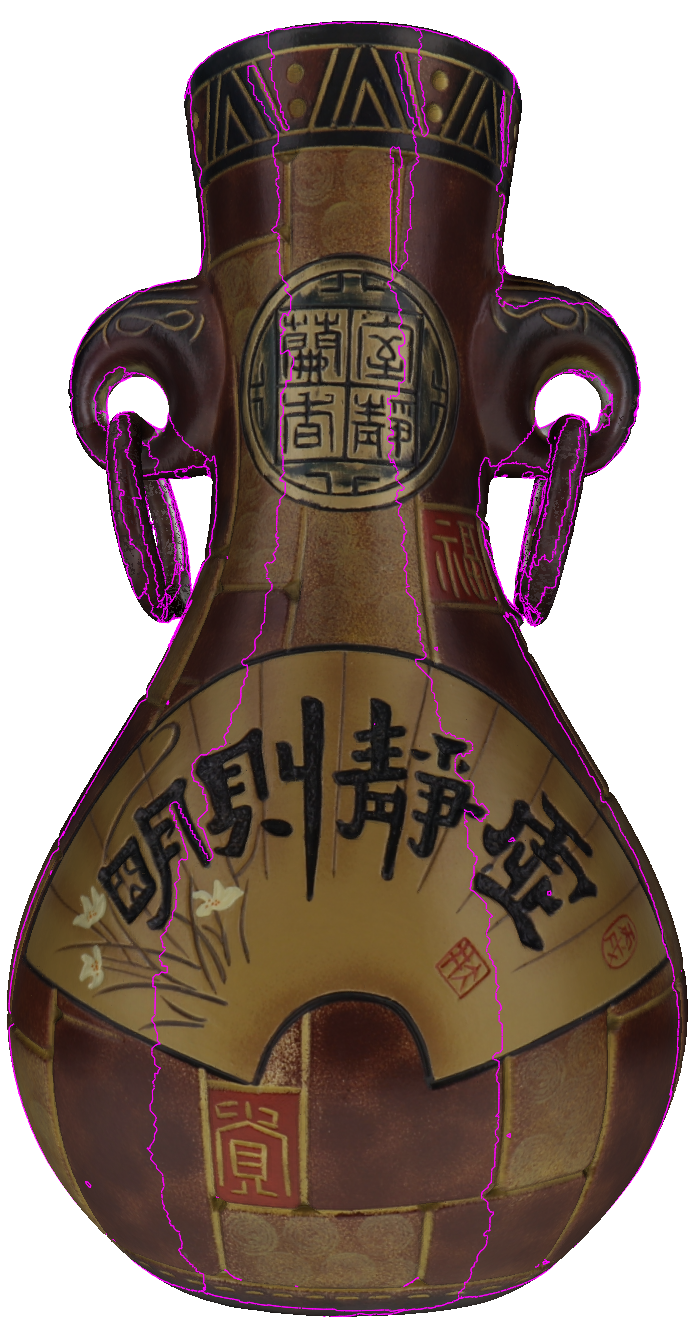}
  }
  \thinspace
  \subcaptionbox{\label{fig:ex5}}{
    \includegraphics[width=0.16\linewidth]{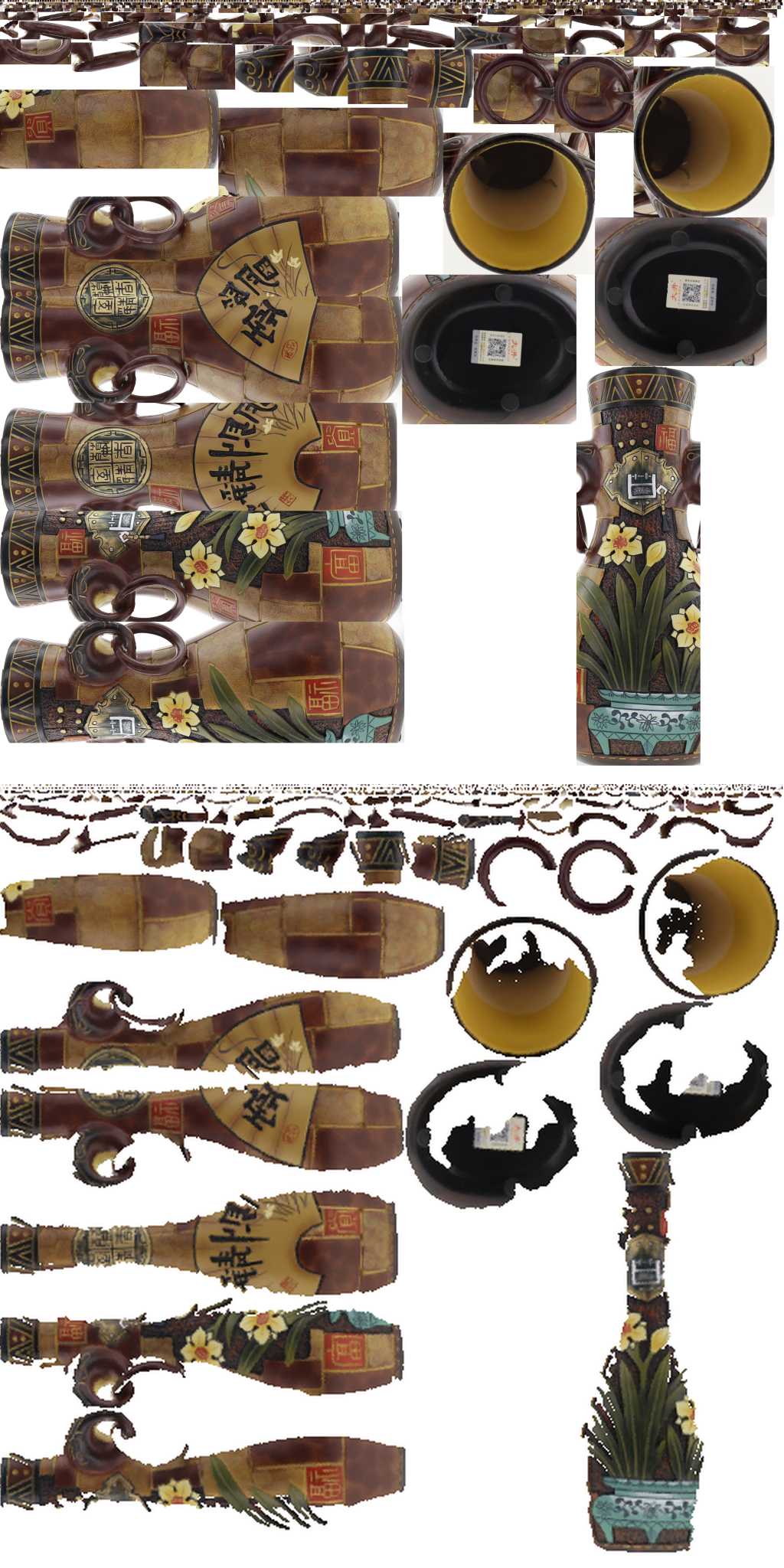}
  }
  \thinspace
  \subcaptionbox{\label{fig:ex6}}{
    \includegraphics[width=0.17\linewidth]{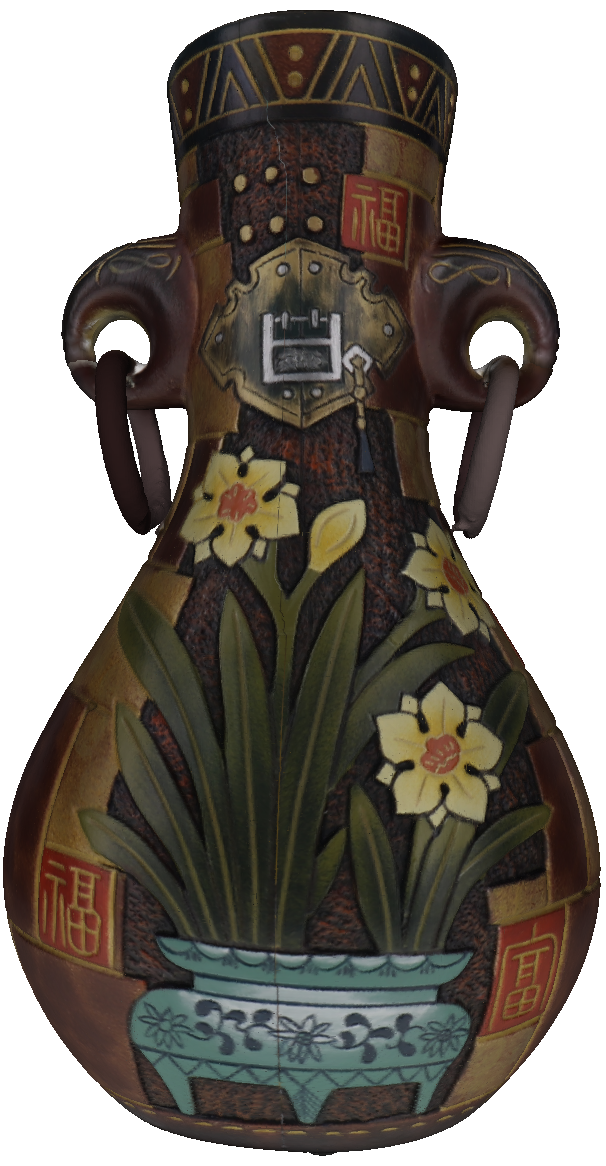}
  }
  \thinspace
  \subcaptionbox{\label{fig:ex7}}{
    \includegraphics[width=0.17\linewidth]{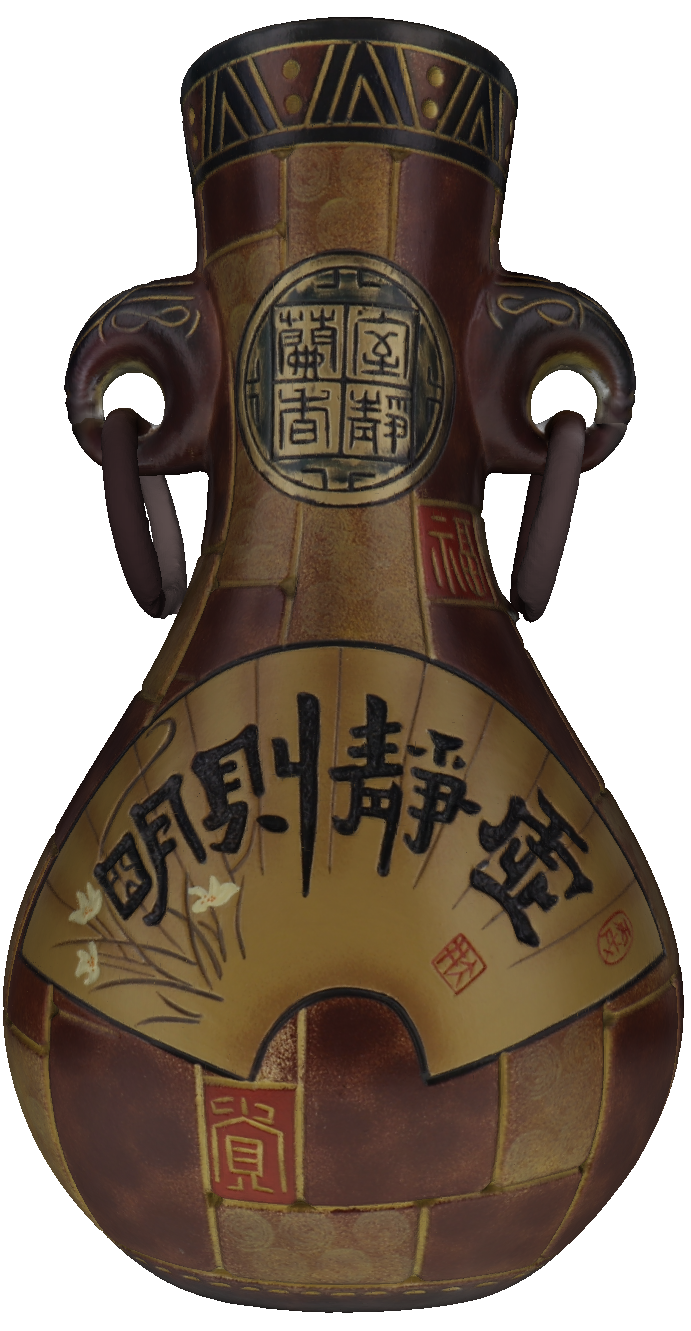}
  }
  \thinspace
  \subcaptionbox{\label{fig:ex8}}{
    \includegraphics[width=0.17\linewidth]{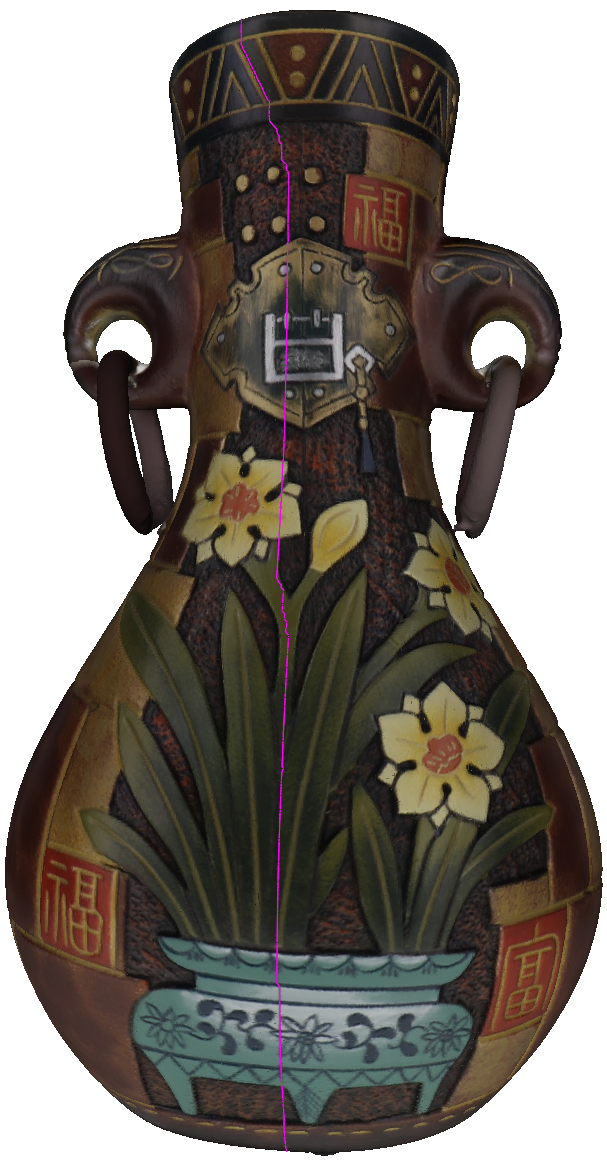}
  }
  \thinspace
  \subcaptionbox{\label{fig:ex9}}{
    \includegraphics[width=0.16\linewidth]{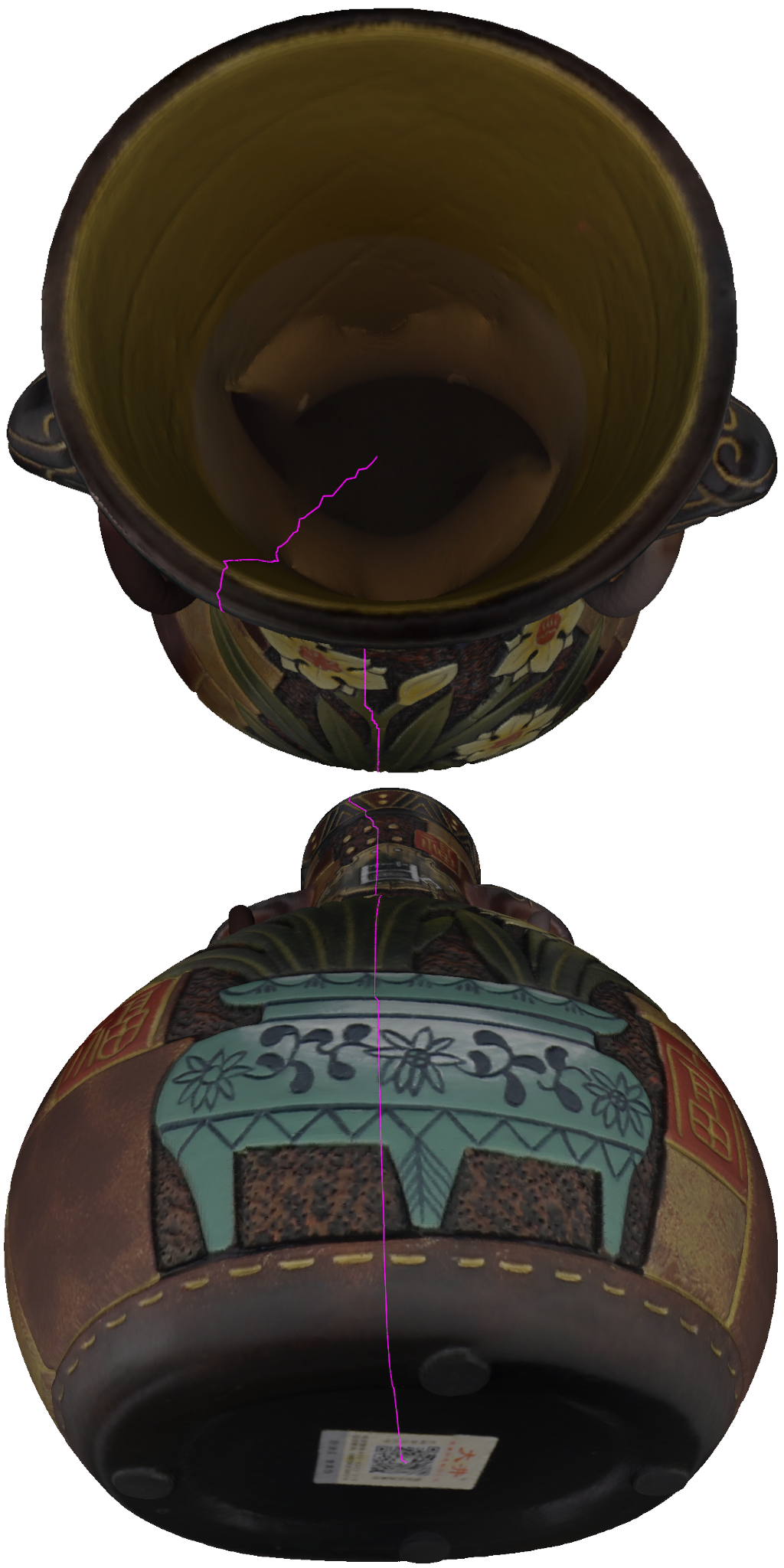}
  }
  \thinspace
  \subcaptionbox{\label{fig:ex10}}{
    \includegraphics[width=0.17\linewidth]{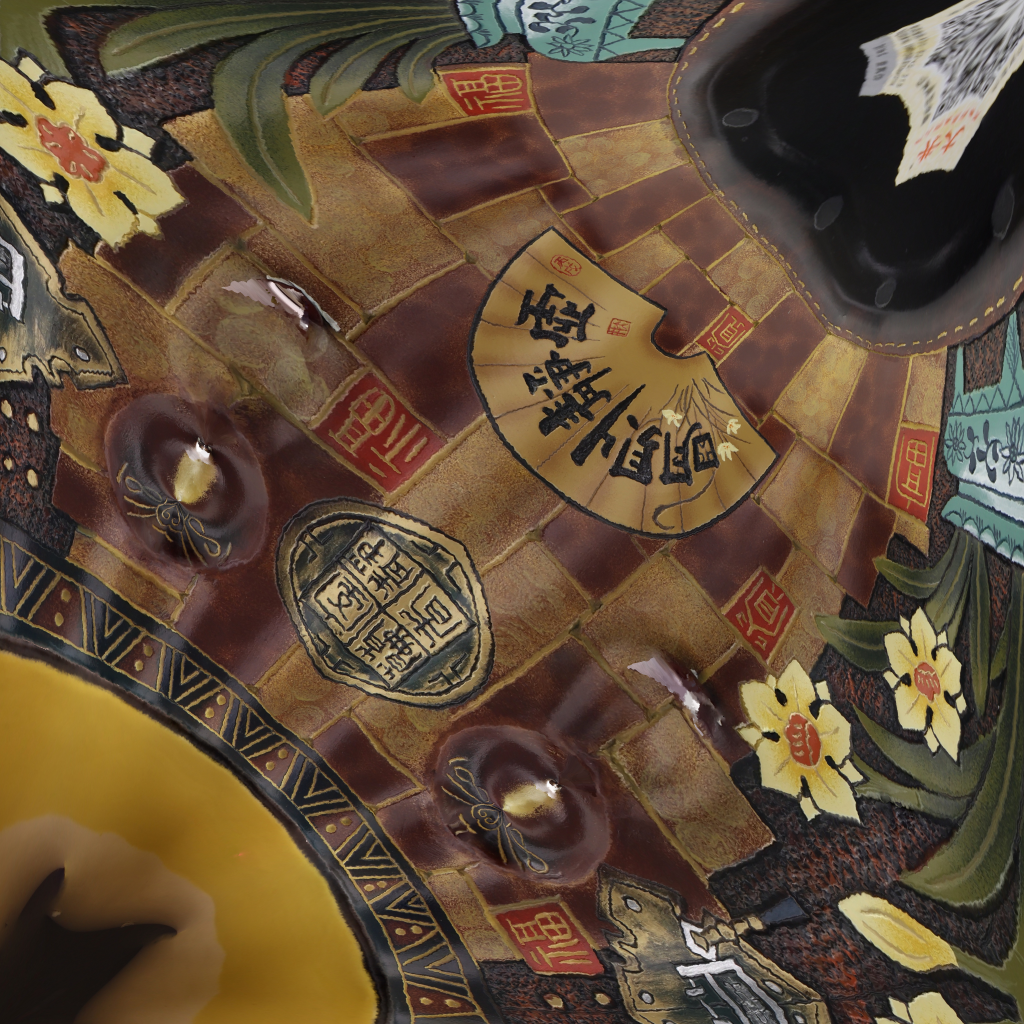}
  }
  \caption{
    Comparison between original (\subref{fig:ex1}-\subref{fig:ex5}) and result vase models (\subref{fig:ex6}-\subref{fig:ex10}):
    \subref{fig:ex1}, \subref{fig:ex2} front and back views of original model, \subref{fig:ex3}, \subref{fig:ex4} the pink lines show the texture border of original model, \subref{fig:ex5} the top one is the input texture (4K resolution), but the other one is the actual texture which used in original model, \subref{fig:ex6}, \subref{fig:ex7} front and back views of result model, visually they look the same as \subref{fig:ex1} and \subref{fig:ex2} respectively, \subref{fig:ex8}, \subref{fig:ex9} there is only one texture border, \subref{fig:ex10} in the result texture (resolution is $4096 \times 4096$ pixels), there is no redundant space or pixel.}
  \label{fig:ex_comparison}
\end{figure}

\cref{fig:ex_comparison} provides a visual comparison between the input vase model and its corresponding output. This model is generated through a structured light scanning system, with a geometric accuracy of approximately 0.5mm. The scanning process involves capturing the vase from various perspectives, reconstructing point clouds using a three-step phase-shifting algorithm~\cite{huang2006fast}. Subsequently, the normal iterative closest point (NICP) algorithm~\cite{besl1992method, serafin2015nicp} is employed for point cloud fusion, resulting in a complete point cloud. The model is then reconstructed by solving a Poisson equation~\cite{kazhdan2006poisson}. Finally, the original texture is generated by projecting multi-view images onto the vase surface. This comprehensive scanning and reconstruction process explains why the original texture border intricately divides the texture space into numerous irregular patches.

Notably, the resulting texture image (\cref{fig:ex10}) seamlessly occupies the entire texture space without any redundant space or pixels. The visual coherence between the input and output is striking. The transformation not only simplifies the texture border but also optimally fills the texture space, showcasing a significant enhancement in visual aesthetics and efficiency.

\begin{figure}[h]
  \centering{}
  \subcaptionbox{\label{fig:ex3_1}}{
    \includegraphics[width=0.22\linewidth]{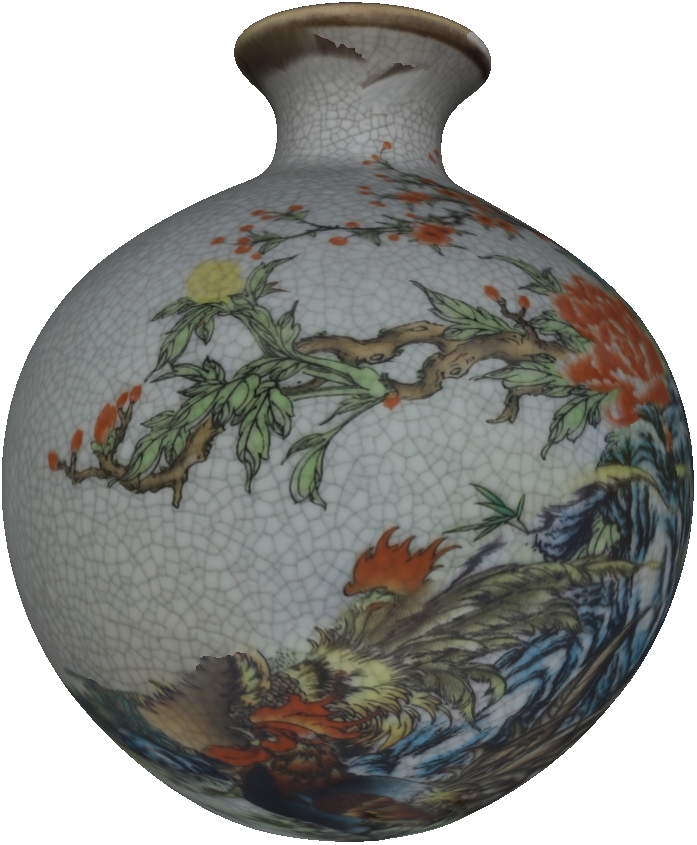}
  }
  \thinspace
  \subcaptionbox{\label{fig:ex3_3}}{
    \includegraphics[width=0.22\linewidth]{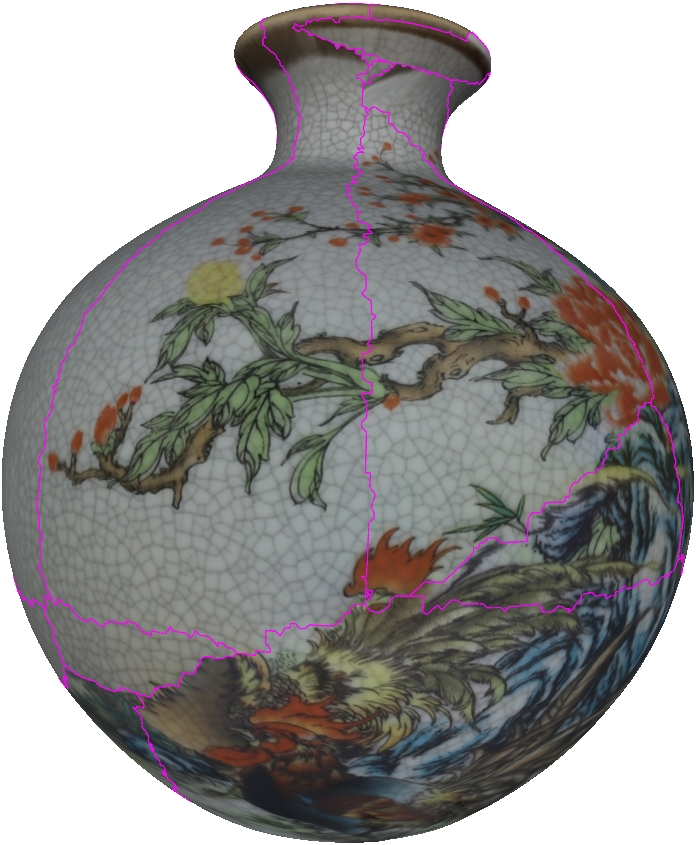}
  }
  \thinspace
  \subcaptionbox{\label{fig:ex3_4}}{
    \includegraphics[width=0.22\linewidth]{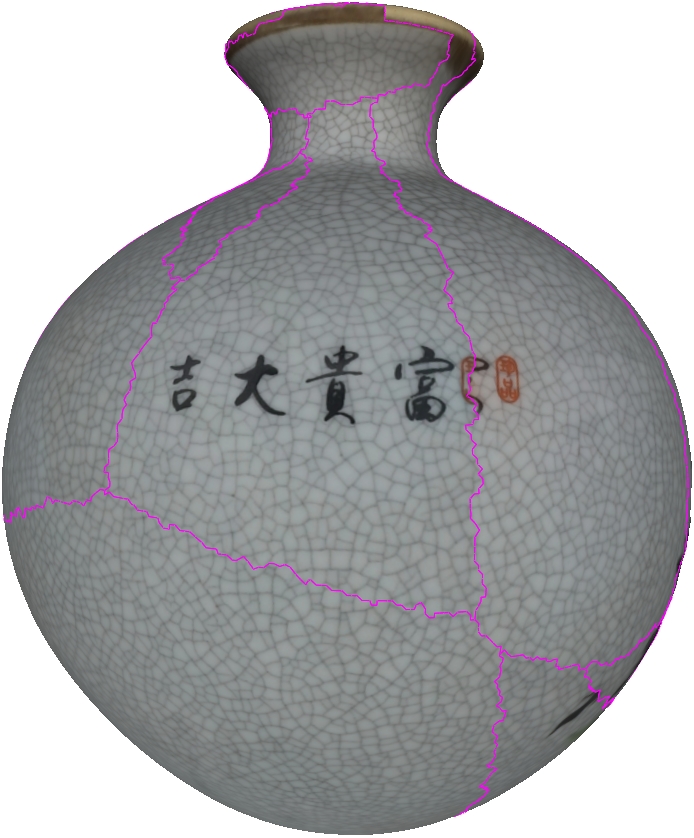}
  }
  \thinspace
  \subcaptionbox{\label{fig:ex3_5}}{
    \includegraphics[width=0.17\linewidth]{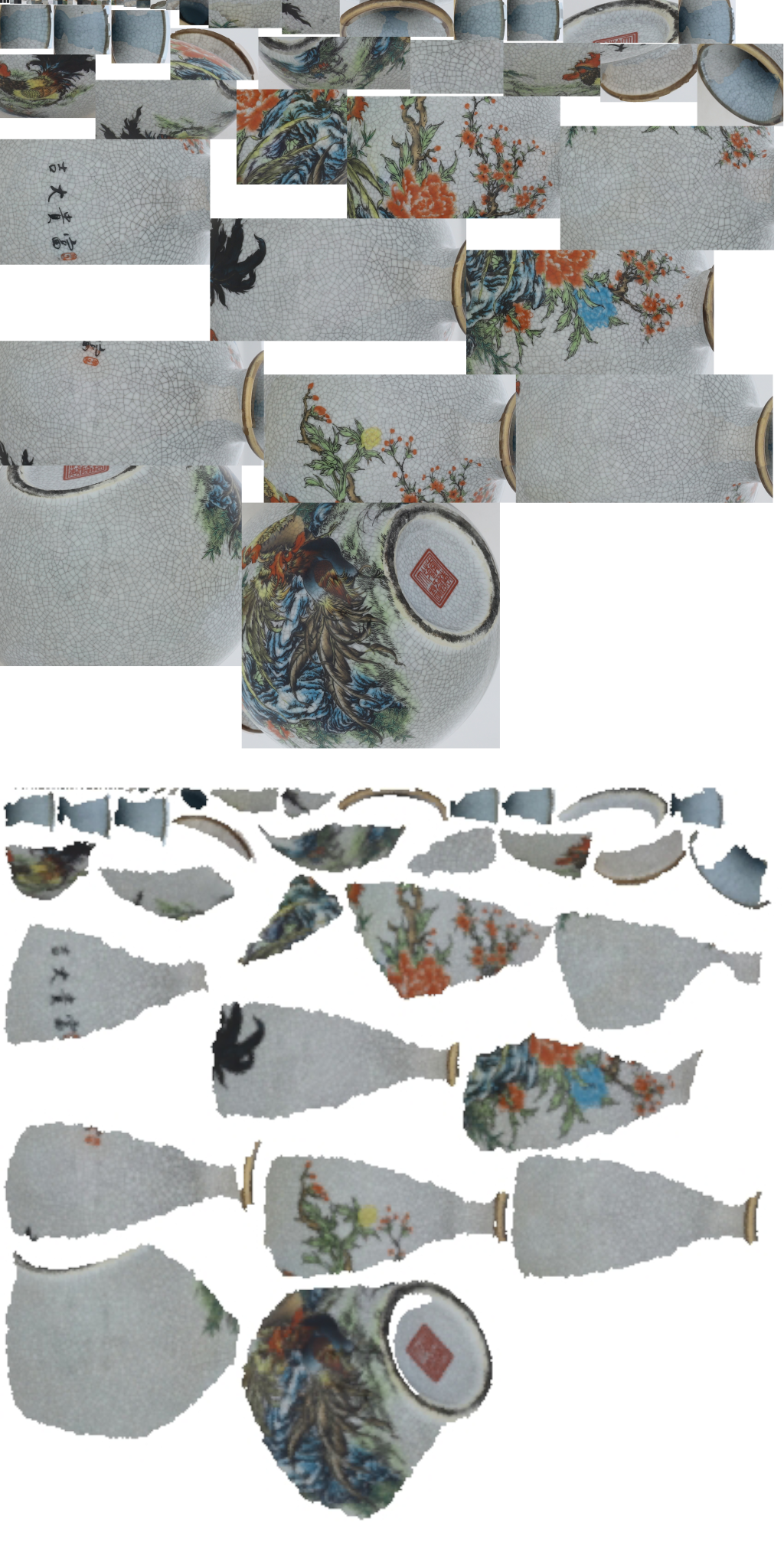}
  }
  \thinspace
  \subcaptionbox{\label{fig:ex3_6}}{
    \includegraphics[width=0.22\linewidth]{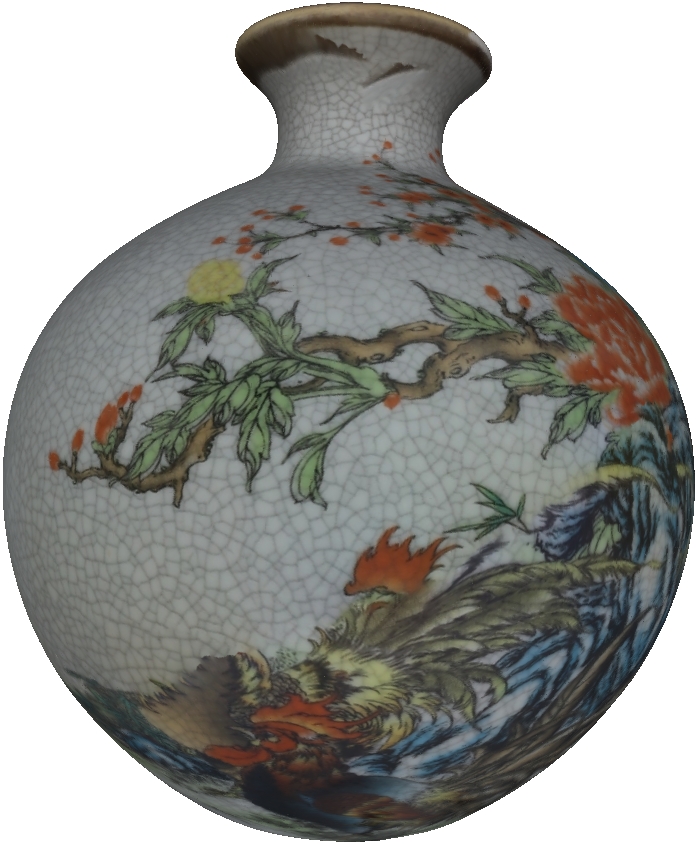}
  }
  \thinspace
  \subcaptionbox{\label{fig:ex3_7}}{
    \includegraphics[width=0.22\linewidth]{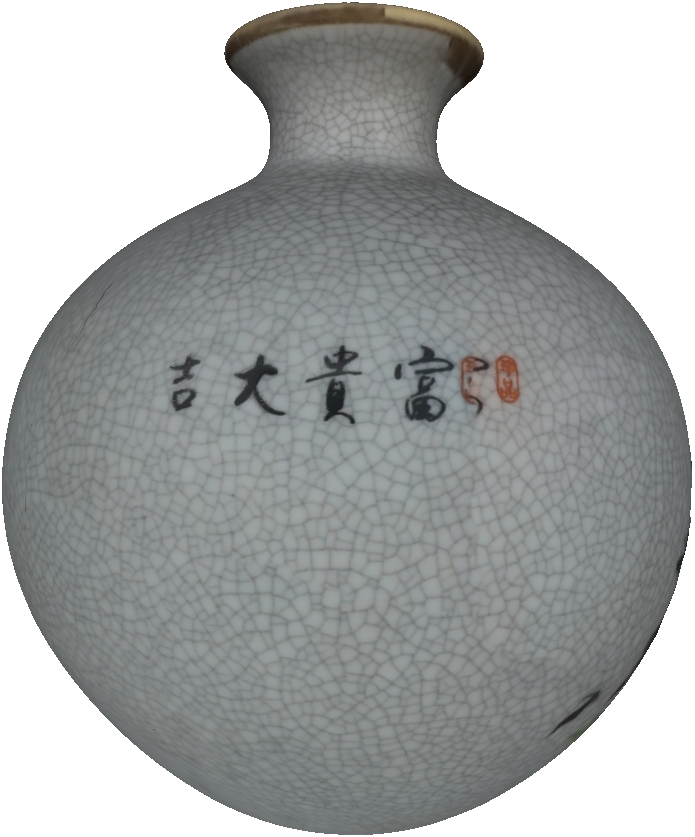}
  }
  \thinspace
  \subcaptionbox{\label{fig:ex3_8}}{
    \includegraphics[width=0.22\linewidth]{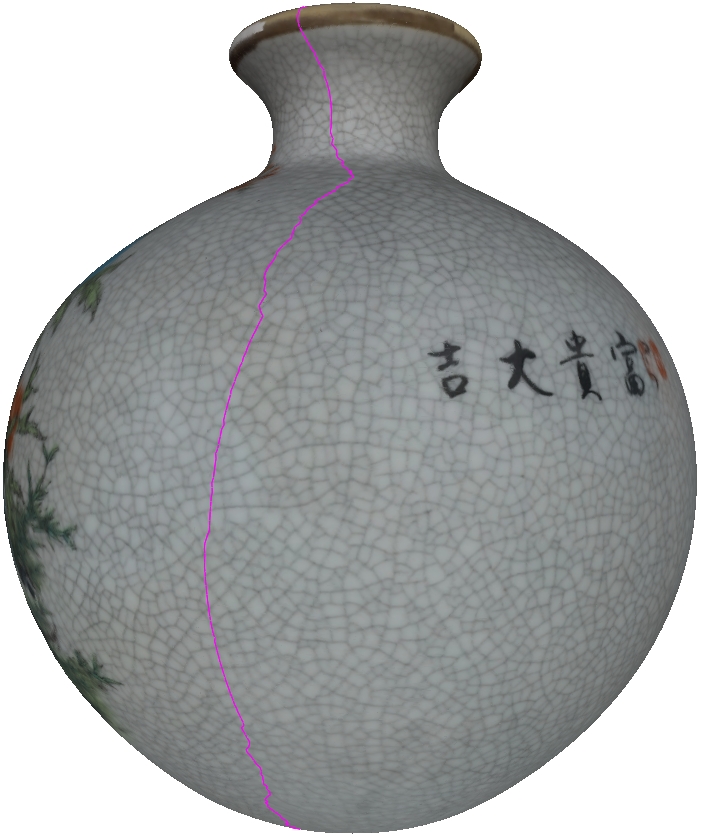}
  }
  \thinspace
  \subcaptionbox{\label{fig:ex3_9}}{
    \includegraphics[width=0.18\linewidth]{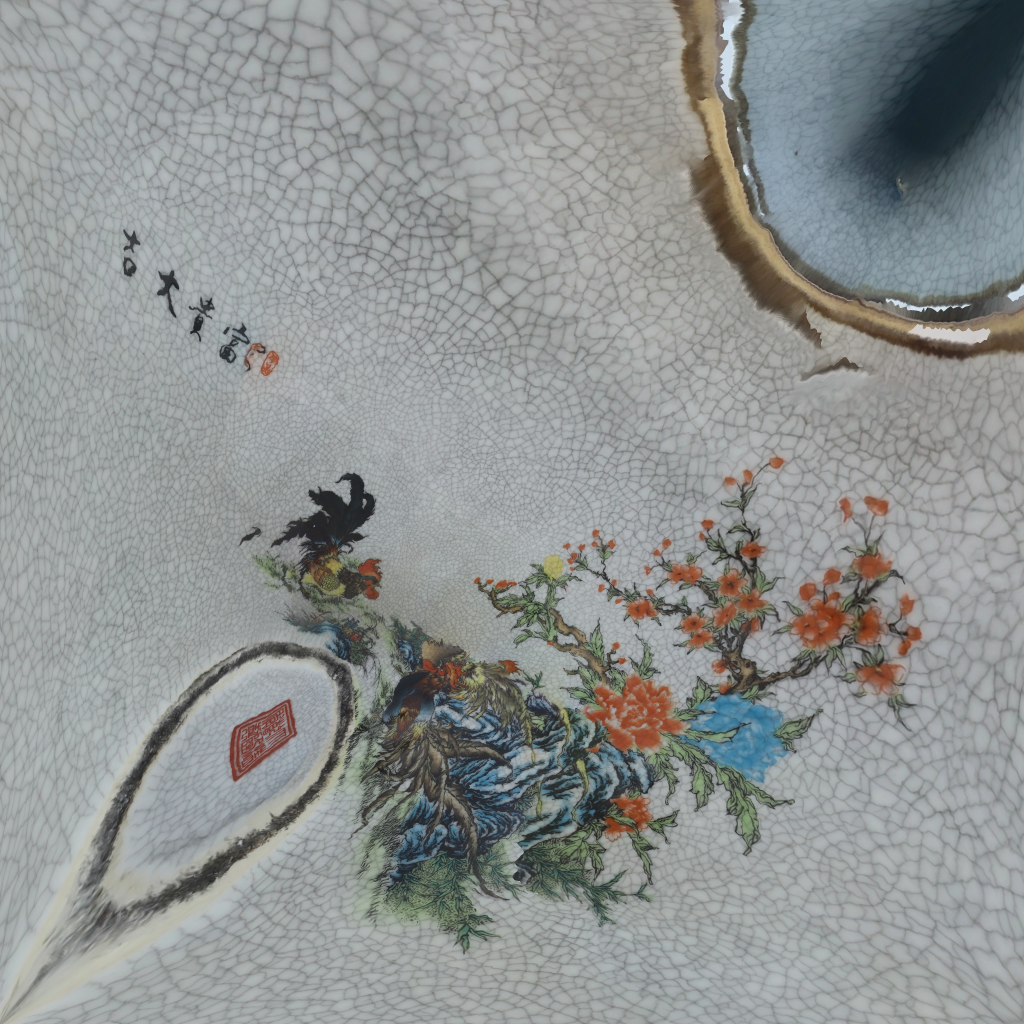}
  }
  \caption{
    Comparison between original (\subref{fig:ex1_1}-\subref{fig:ex1_4}) and result vase models (\subref{fig:ex1_5}-\subref{fig:ex1_9}):
    \subref{fig:ex1_1} front view of original model, \subref{fig:ex1_2}, \subref{fig:ex1_3} the pink lines show the texture border of original model, \subref{fig:ex1_4} the top one is the input texture (resolution is $4711 \times 4711$ pixels), but the other one is the actual texture which used in original model, \subref{fig:ex1_5}, \subref{fig:ex1_6} front and back views of result model, visually they look the same as the input model, \subref{fig:ex1_7} there is only one texture border, \subref{fig:ex1_9} the output texture (resolution is $4096 \times 4096$ pixels).}
  \label{fig:ex_comparison3}
\end{figure}

\cref{fig:ex_comparison3} presents a comparative analysis between the input vase model and the generated result. The model is also created by a structured light scanning system, achieving a geometric accuracy of approximately 0.5mm. In the original model, the texture border intricately divides the texture space into numerous irregular patches. In contrast, the resultant texture border forms a simple loop, traversing from one endpoint to the other and backtracking to the source. Remarkably, the output texture image (\cref{fig:ex3_9}) effectively occupies the entire texture space without any redundant space or pixels. The visual representation of the input and output appears consistent.

\begin{figure}[h]
  \centering{}
  \subcaptionbox{\label{fig:ex1_1}}{
    \includegraphics[width=0.2\linewidth]{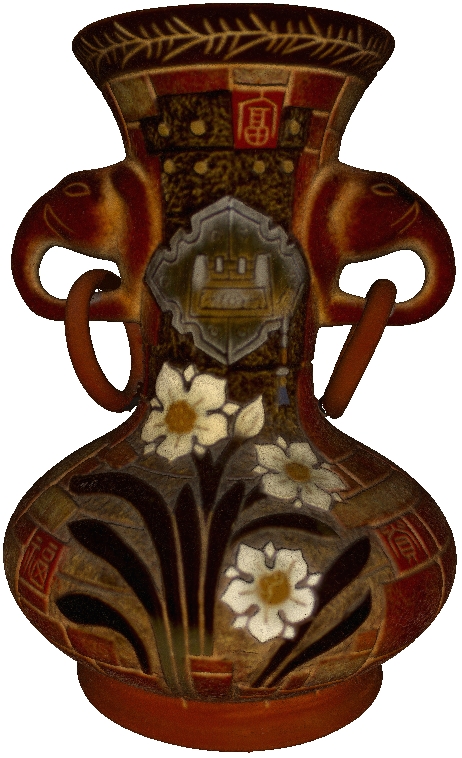}
  }
  \thinspace
  \subcaptionbox{\label{fig:ex1_2}}{
    \includegraphics[width=0.2\linewidth]{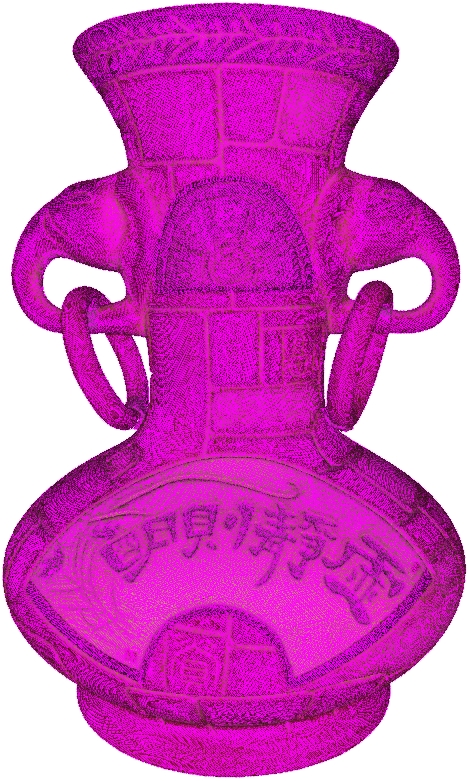}
  }
  \thinspace
  \subcaptionbox{\label{fig:ex1_3}}{
    \includegraphics[width=0.16\linewidth]{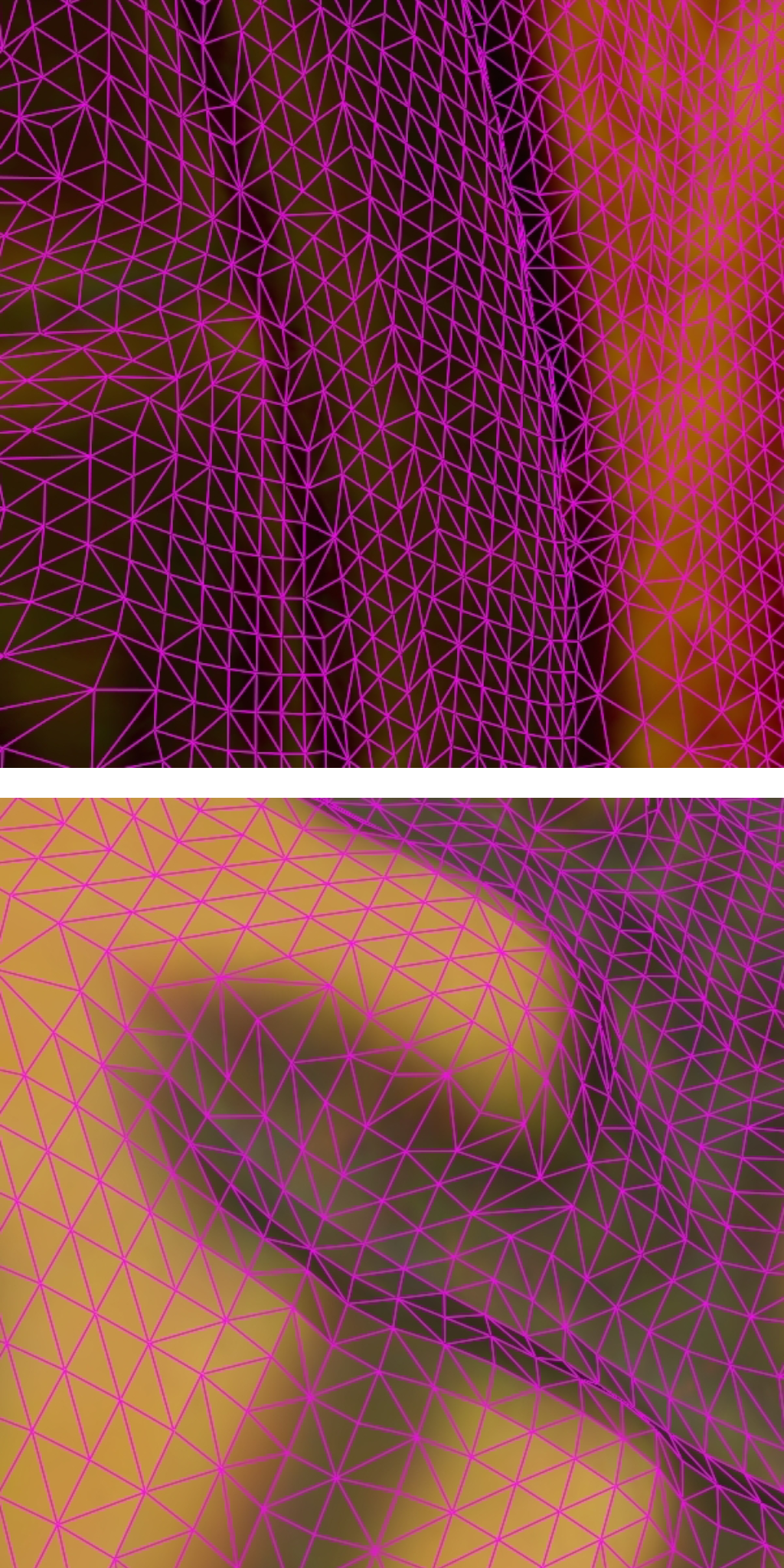}
  }
  \thinspace
  \subcaptionbox{\label{fig:ex1_4}}{
    \includegraphics[width=0.16\linewidth]{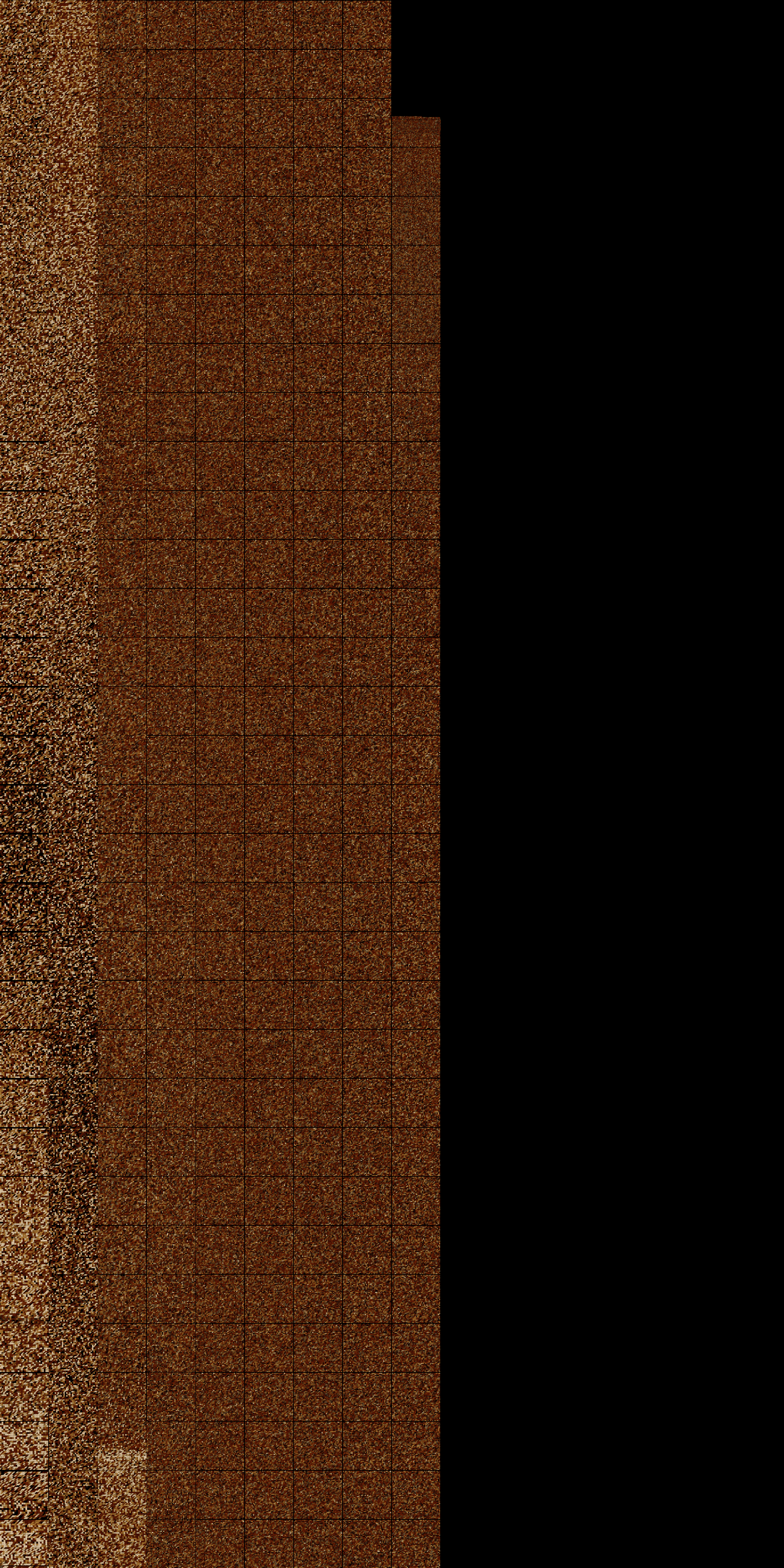}
  }
  \thinspace
  \subcaptionbox{\label{fig:ex1_5}}{
    \includegraphics[width=0.16\linewidth]{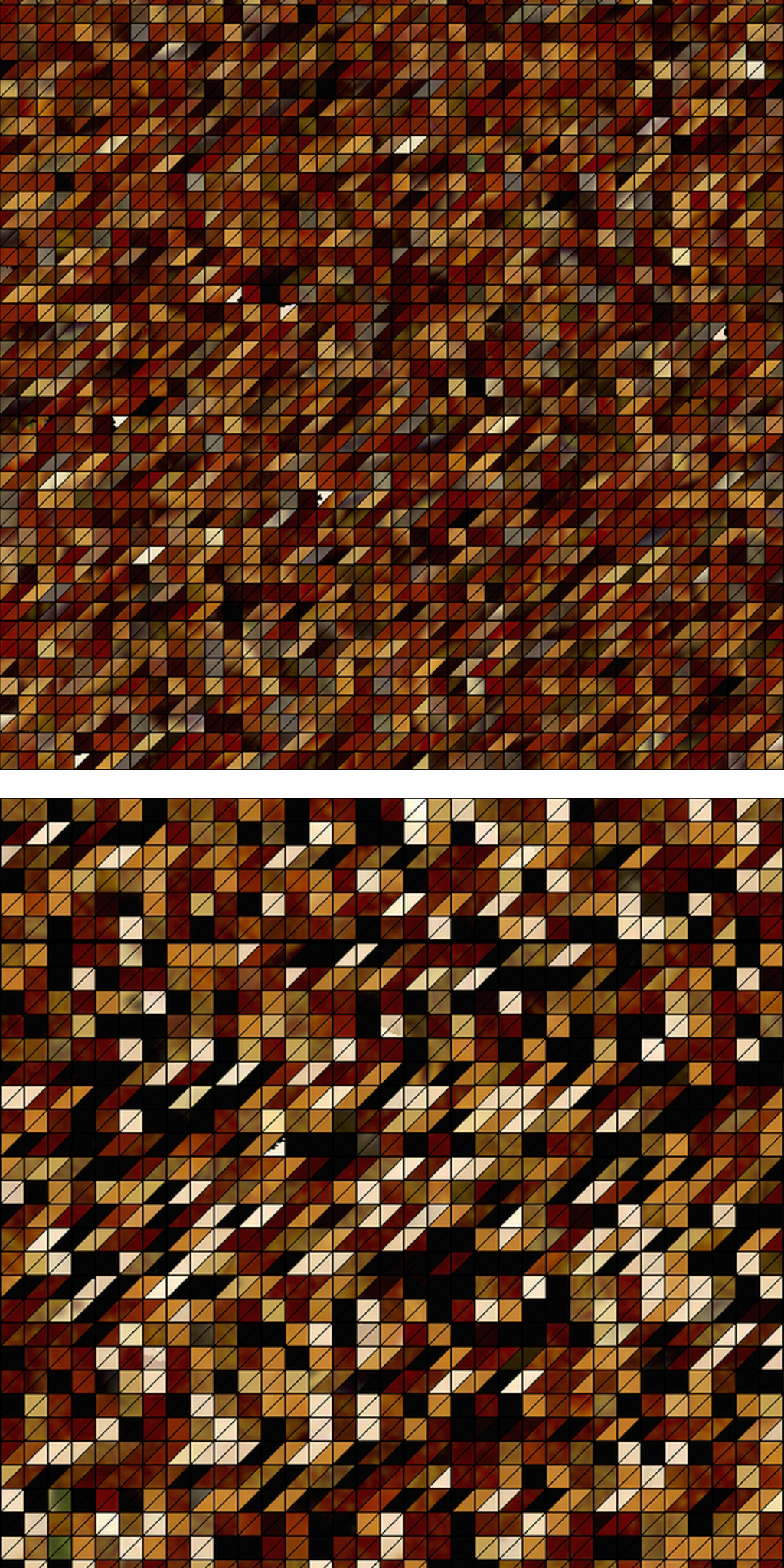}
  }
  \thinspace
  \subcaptionbox{\label{fig:ex1_6}}{
    \includegraphics[width=0.2\linewidth]{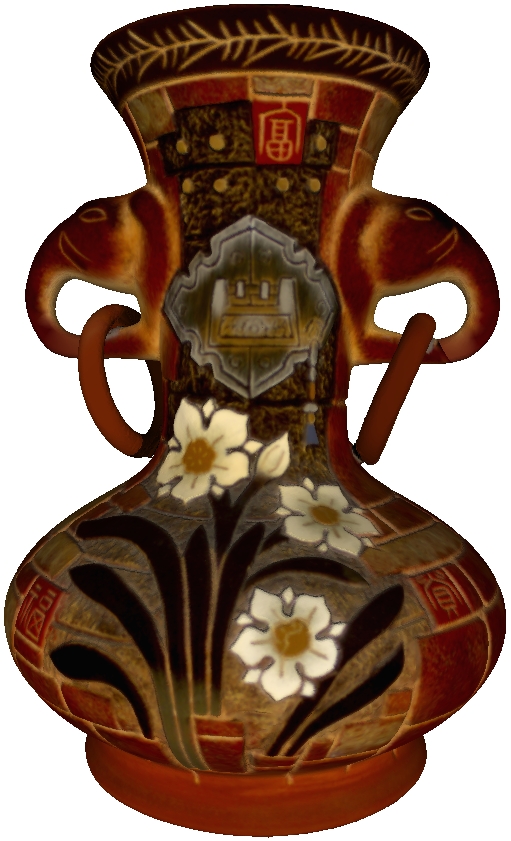}
  }
  \thinspace
  \subcaptionbox{\label{fig:ex1_7}}{
    \includegraphics[width=0.192\linewidth]{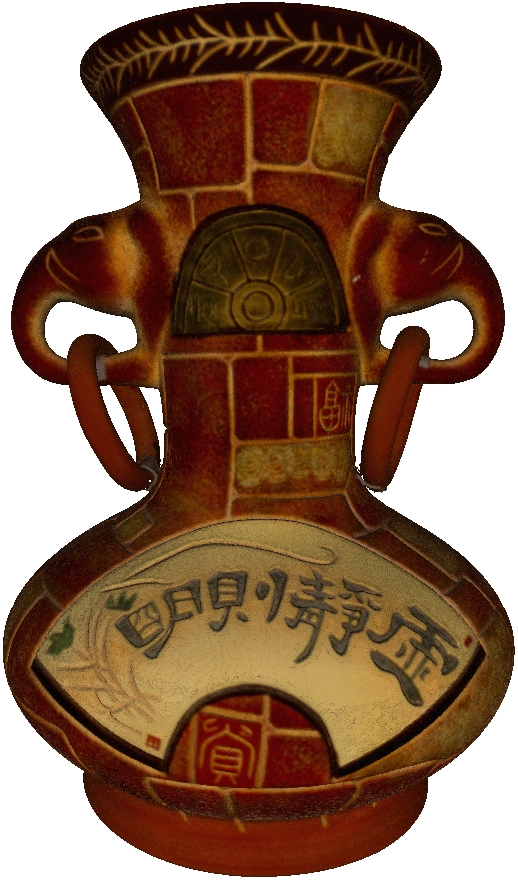}
  }
  \thinspace
  \subcaptionbox{\label{fig:ex1_8}}{
    \includegraphics[width=0.2\linewidth]{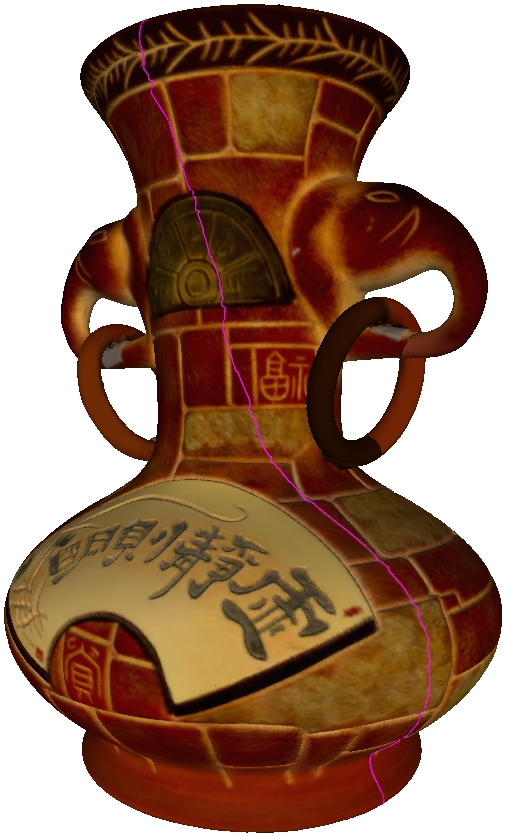}
  }
  \thinspace
  \subcaptionbox{\label{fig:ex1_9}}{
    \includegraphics[width=0.31\linewidth]{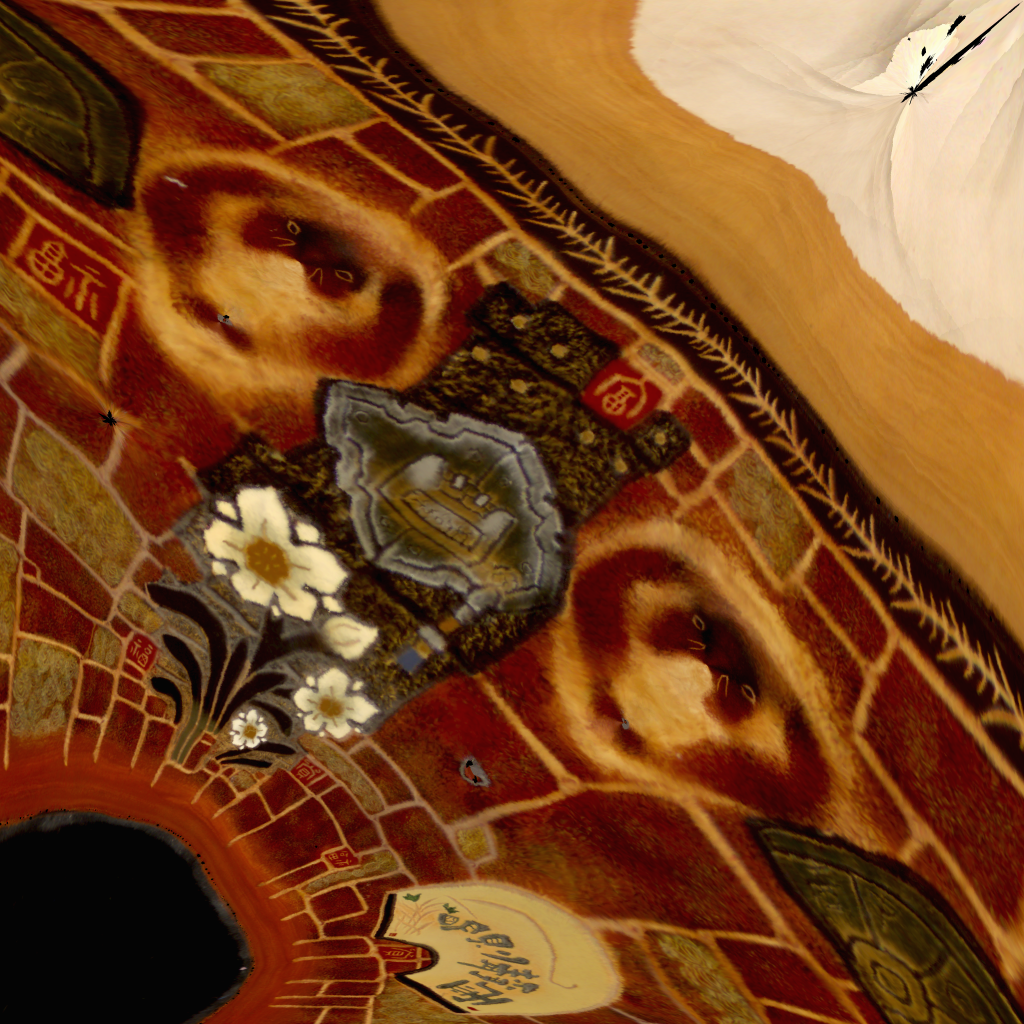}
  }
  \caption{
    Comparison between original (\subref{fig:ex1_1}-\subref{fig:ex1_5}) and output vase models (\subref{fig:ex1_6}-\subref{fig:ex1_9}):
    \subref{fig:ex1_1} the front view of original model, \subref{fig:ex1_2} in this case, each triangle of the original model has a distinct texture space, and \subref{fig:ex1_3} shows two enlarged parts, \subref{fig:ex1_4} the input texture (resolution is $8192 \times 16384$ pixels), from the enlarged parts \subref{fig:ex1_5} we can find that all texture spaces are triangles, \subref{fig:ex1_6}, \subref{fig:ex1_7} the front and back views of result model, \subref{fig:ex1_8} the texture border is reduced to a simple one, \subref{fig:ex1_9} the result texture (resolution is $4096 \times 4096$ pixels) has no redundant space or pixel, and looks more unified.}
  \label{fig:ex_comparison1}
\end{figure}

\cref{fig:ex_comparison1} illustrates an extreme case where the initial texture of the input vase model is formed by individual triangles, each with its own distinct texture space. Upon employing our method, the resultant texture border is streamlined into a cohesive loop. The output texture image appears more unified, showcasing the effectiveness of our approach in enhancing texture space optimization.

\begin{figure}[!h]
  \centering{}
  \subcaptionbox{\label{fig:ex2_1}}{
    \includegraphics[width=0.48\linewidth]{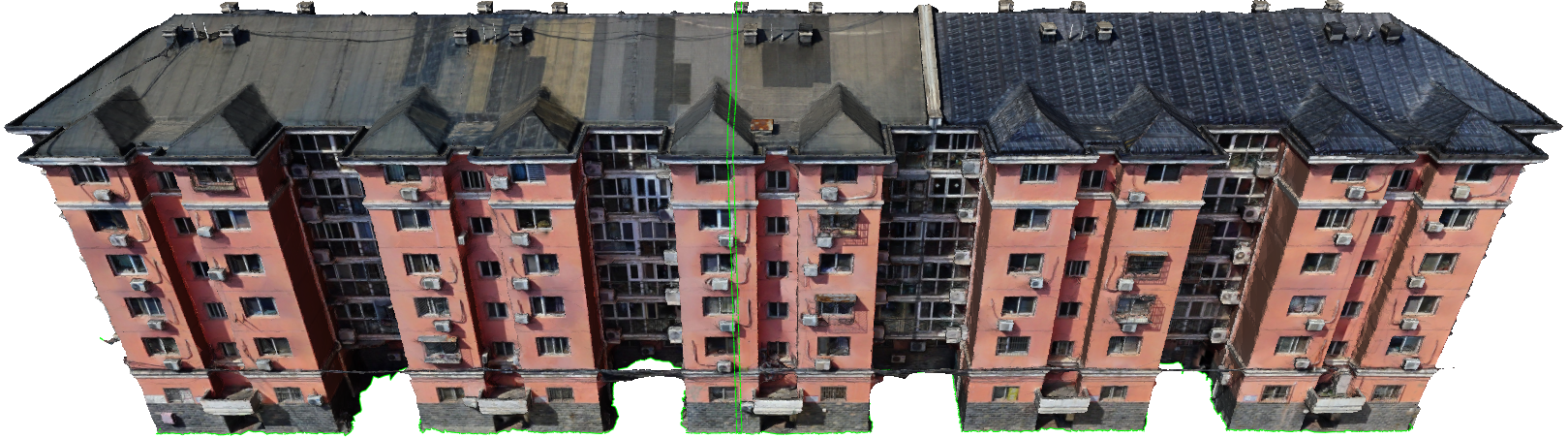}
  }
  \thinspace
  \subcaptionbox{\label{fig:ex2_2}}{
    \includegraphics[width=0.48\linewidth]{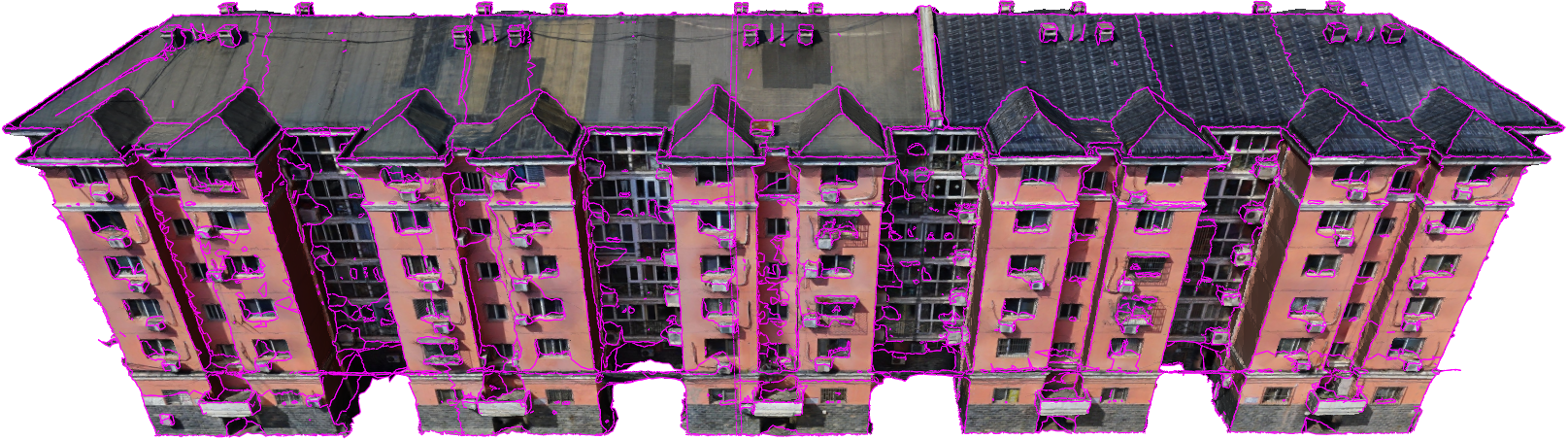}
  }
  \thinspace
  \subcaptionbox{\label{fig:ex2_3}}{
    \includegraphics[width=0.48\linewidth]{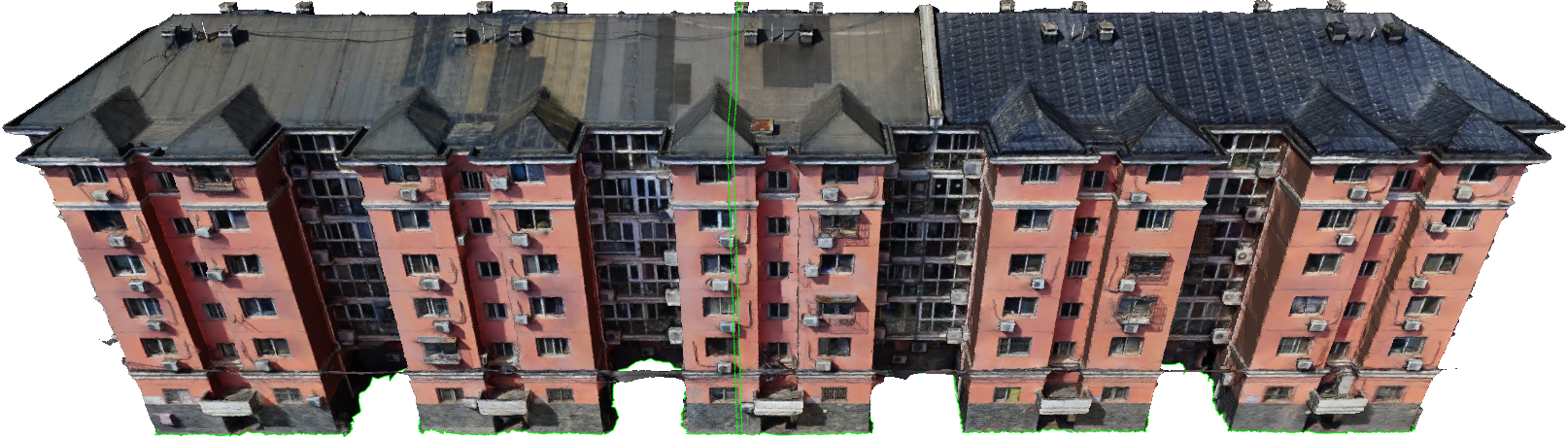}
  }
  \thinspace
  \subcaptionbox{\label{fig:ex2_4}}{
    \includegraphics[width=0.48\linewidth]{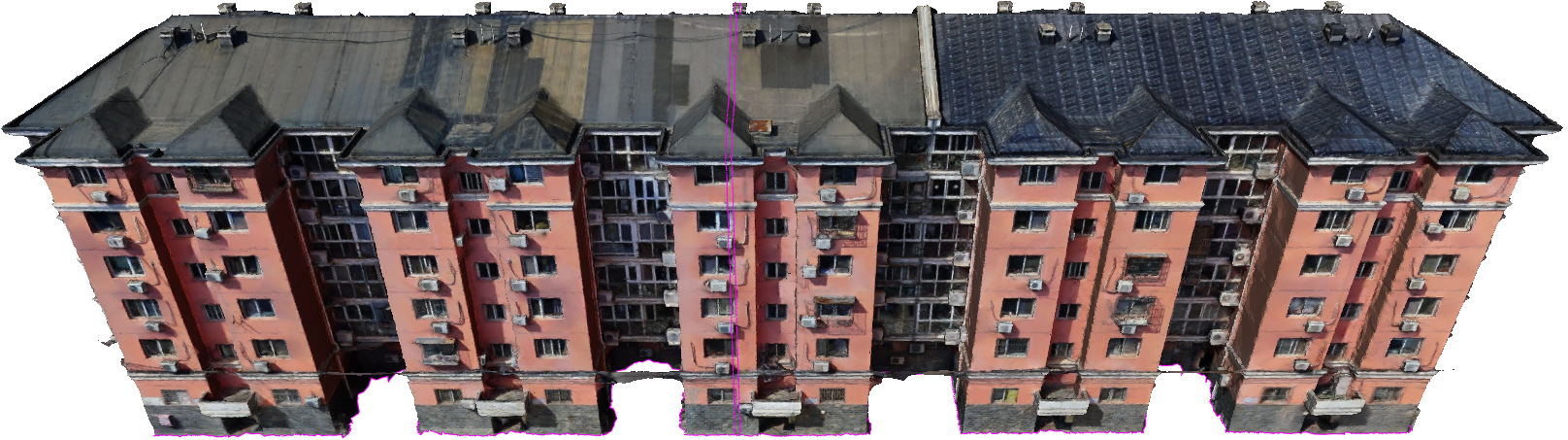}
  }
  \thinspace
  \subcaptionbox{\label{fig:ex2_5}}{
    \includegraphics[width=0.48\linewidth]{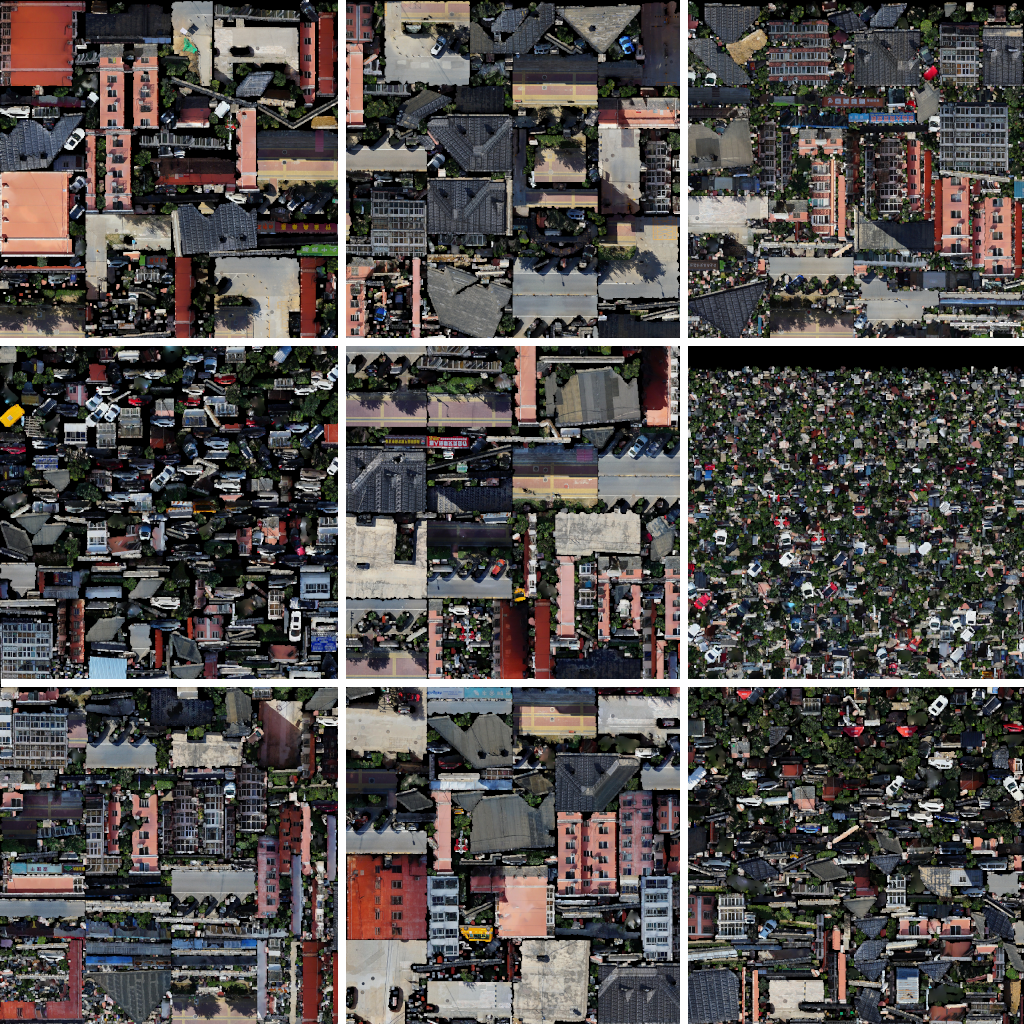}
  }
  \thinspace
  \subcaptionbox{\label{fig:ex2_6}}{
    \includegraphics[width=0.16\linewidth]{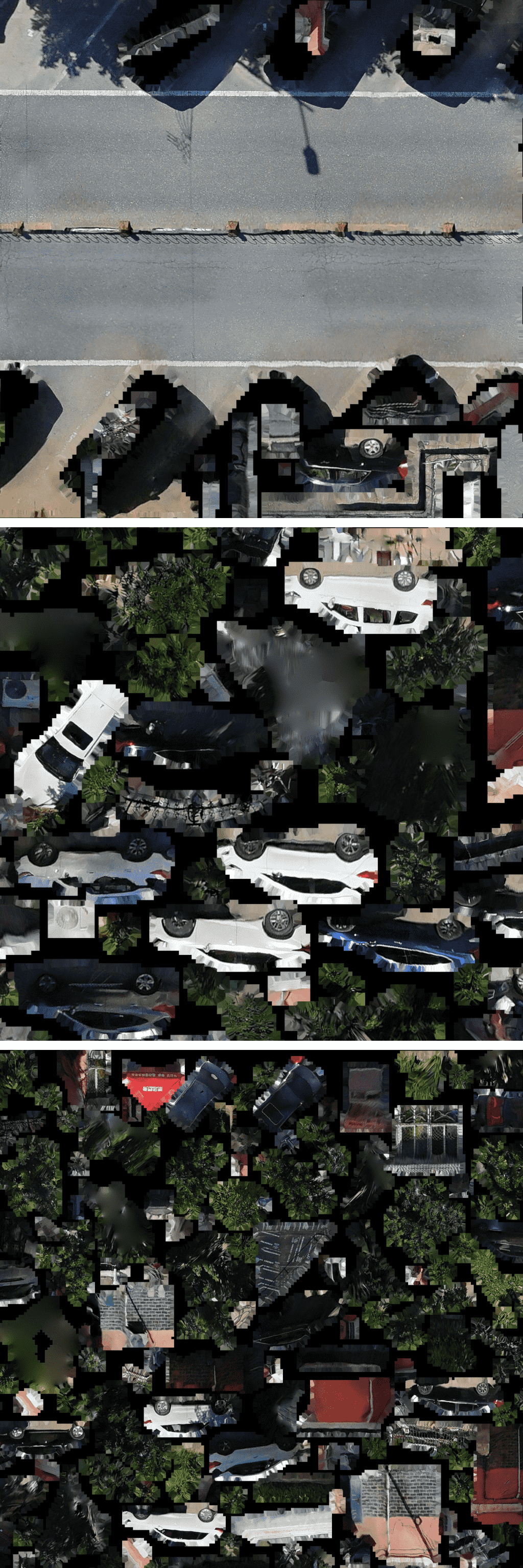}
  }
  \thinspace
  \subcaptionbox{\label{fig:ex2_7}}{
    \includegraphics[width=0.24\linewidth]{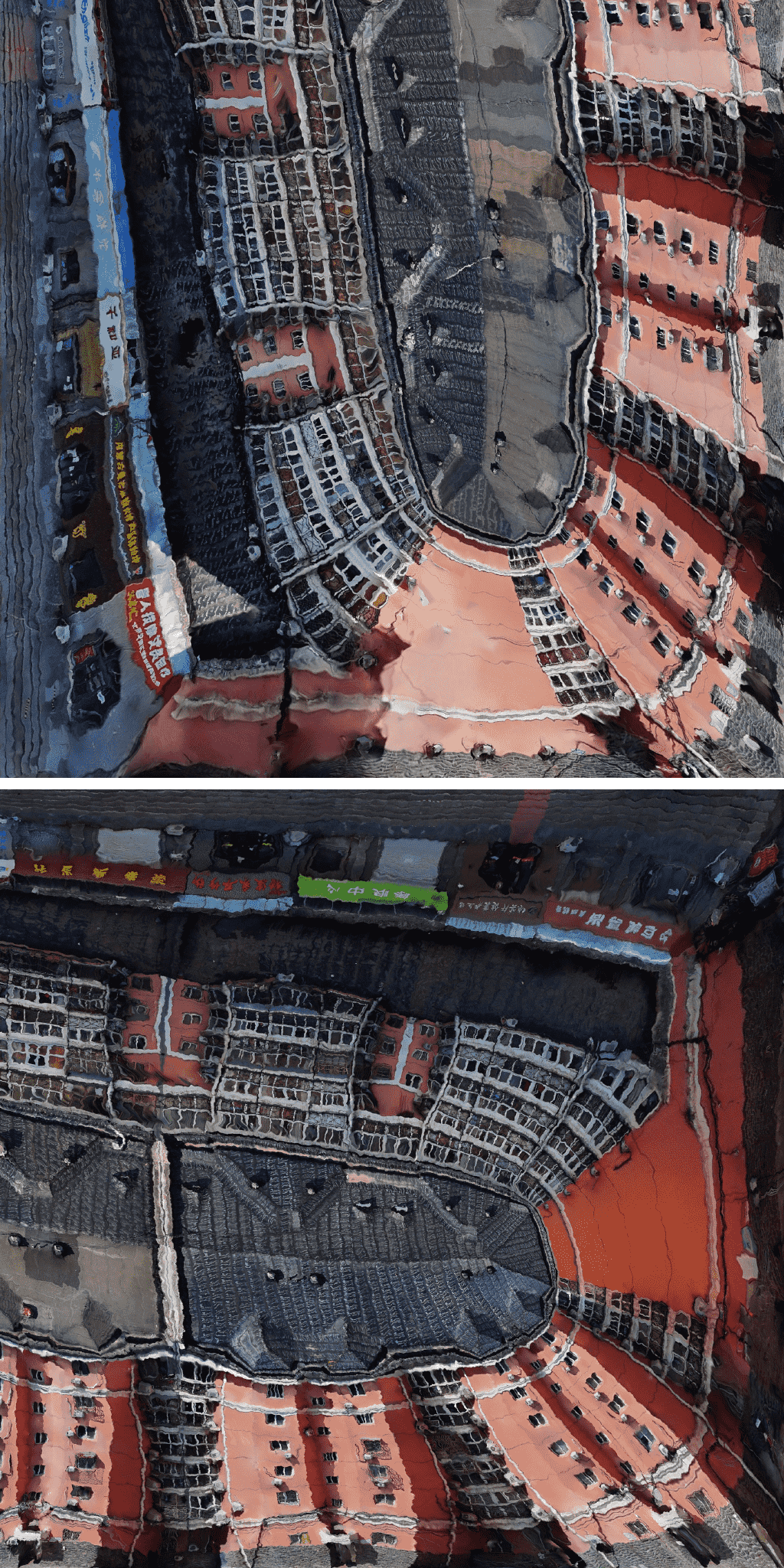}
  }
  \caption{
    Comparison between original (\subref{fig:ex2_1},\subref{fig:ex2_2},\subref{fig:ex2_5} and ,\subref{fig:ex2_6}) and result building models (\subref{fig:ex2_3}, ,\subref{fig:ex2_4} and \subref{fig:ex2_7}):
    \subref{fig:ex2_1} the model has two parts, the green lines show the surface boundaries of the two parts, \subref{fig:ex2_2} the pink lines show the texture border of the building model, \subref{fig:ex2_3} the result model which owns the same geometry and topology, \subref{fig:ex2_4} the texture border simply separates the texture space into two subspaces, \subref{fig:ex2_5} contains nine texture images (each one is 4K resolution) of the original model, \subref{fig:ex2_6} shows that there are lots of irrelevant pixels exist in \subref{fig:ex2_5}, \subref{fig:ex2_7} the two result textures (4K resolution), there is no redundant space or pixel.}
  \label{fig:ex_comparison2}
\end{figure}

\cref{fig:ex_comparison2} shows another experiment of a building reconstructed by multi-view stereo method. The model has two components, so we process these parts according to the pipeline in~\cref{fig:pipeline} respectively. From~\cref{fig:ex2_2}, we can observe that the texture space is messy and fragmented, and in~\cref{fig:ex2_4} the proposed method improves the situation heavily. There is no redundancy space but there exists many irrelevant pixels in the original textures (\cref{fig:ex2_5}), such as roads, cars, trees and even other buildings (\cref{fig:ex2_6}). In \cref{fig:ex2_7}, we generate two textures with resolution of $4096 \times 4096$ pixels for the new texture coordinates.

In the aforementioned experiments, the newly generated textures demonstrate the substantial optimization achieved by our method in terms of texture space utilization. This optimization involves the elimination of unused pixels and the efficient utilization of the available texture space, leading to streamlined data storage and enhancing GPU rendering efficiency.

However, it's important to acknowledge a drawback associated with our proposed method. The use of harmonic map in the global parameterization may introduce area distortions, causing certain regions of the generated texture to shrink significantly. This distortion can result in a loss of texture quality in those specific areas.

\section{Conclusions}
\label{sec:conclusions}
In this work, we propose a novel texture space optimization method. The proposed method is based on harmonic map which is an algorithm of global parameterization. The algorithms and technology details in this proposed method are described. Experimental results show that our method can generate textures without redundancy space and pixels. As a result, this method will save both data storage and GPU memory addressing time in the rendering process, making the method amenable for storage-critical applications in computer graphics and other fields. However, the harmonic mapping introduces area distortion in the result texture, this may weaken the quality of the texture images. For future work, we will improve the parameterization to achieve more efficient data storage for texture space optimization task.

\section*{Acknowledgments}
\label{sec:acknowledgments}
This research was supported by the National Natural Science Foundation of China T2225012,  61936002, and the Fundamental Research Funds for the Central Universities DUT22QN212.

\clearpage
\bibliographystyle{abbrv}
\bibliography{ref}

\end{document}